%% file: main.tex
\documentclass[runningheads]{llncs}

\usepackage{eccv}

\usepackage{eccvabbrv}

\usepackage{graphicx}
\usepackage{booktabs}

\newcommand{\vect}[1]{\mbox{\boldmath{$#1$}}}
\definecolor{shq}{rgb}{0.9,0.1,0.1}

\usepackage[accsupp]{axessibility}  % Improves PDF readability for those with disabilities.

\usepackage[pagebackref,breaklinks,colorlinks,citecolor=eccvblue]{hyperref}
\usepackage{orcidlink}
\def\ourmethod{{$D^{2}RF$}\xspace}
\usepackage{nicematrix}
\usepackage{nicematrix}
\usepackage{tabularx}
\usepackage{multirow}
\usepackage{enumitem}
\usepackage{threeparttable}
\usepackage{eucal}
\usepackage{ragged2e}
\usepackage{algorithm}
\usepackage{algorithmic}
\usepackage{makecell}
\usepackage{amssymb}
\definecolor{shq}{rgb}{0.1,0.1,0.9}

\begin{document}

% \title{Layered Depth-of-Field Meets \\ Dynamic Neural Radiance Fields} 

% \title{Supplementary Material for \\ Dynamic Neural Radiance Field\\ From Defocused Monocular Video} 
\title{Dynamic Neural Radiance Field\\ From Defocused Monocular Video} 


\author{Xianrui Luo\inst{1}\orcidlink{0000-0001-8572-8938} \and
Huiqiang Sun\inst{1}\orcidlink{0000-0002-3653-3613} \and
Juewen Peng\inst{2}\orcidlink{0000-0001-5740-2682}\and
Zhiguo Cao\inst{1}\orcidlink{0000-0002-9223-1863}}

\authorrunning{X. Luo et al.}

\institute{Key Laboratory of Image Processing and Intelligent Control, Ministry of Education,\\School of AIA, Huazhong University of Science and Technology, China
\email{\{xianruiluo,shq1031,zgcao\}@hust.edu.cn}\\
\and
College of Computing and Data Science, Nanyang Technological University, Singapore\\
\email{juewen.peng@ntu.edu.sg} \\
\url{https://github.com/xianrui-luo/D2RF}
}
\maketitle

% \begin{abstract}
%   This is required for papers in LNCS proceedings.
%   \keywords{Novel View Synthesis \and Neural Radiance Field \and Depth-of-Field}
% \end{abstract}

\input{sec/0_abstract}    
\input{sec/1_intro}
\input{sec/2_related}

\input{sec/3_method}
\input{sec/4_exp}

\input{sec/5_conclusion}
% \input{sec/supp}

% \clearpage\mbox{}Page \thepage\ of the manuscript. This is the last page.
% Now we have reached the maximum length of an ECCV \ECCVyear{} submission (excluding references).
% References should start immediately after the main text, but can continue past p.\ 14 if needed.
% \clearpage  % TODO REVIEW/FINAL: This \clearpage needs to be removed from both review and camera-ready versions.

\bibliographystyle{splncs04}
\bibliography{main}

\newpage
\input{sec/supp}
\end{document}

%% file: sec/0_abstract.tex
\begin{abstract}
Dynamic Neural Radiance Field (NeRF) from monocular videos has recently been explored for space-time novel view synthesis and achieved excellent results. 
However, defocus blur caused by depth variation often occurs in video capture, compromising the quality of dynamic reconstruction because the lack of sharp details interferes with modeling temporal consistency between input views. 
% However, defocus blur caused by depth variation often occurs in video capture, compromising the quality of dynamic reconstruction because the lack of sharp details interferes with reconstruction and modeling temporal consistency between input views. 
To tackle this issue, we propose \ourmethod, the first dynamic NeRF method designed to restore sharp novel views from defocused monocular videos. 
We introduce layered Depth-of-Field (DoF) volume rendering to model the defocus blur and reconstruct a sharp NeRF supervised by defocused views. 
The blur model is inspired by the connection between DoF rendering and volume rendering. 
% The blur modeling is inspired by the connection between layered Depth-of-Field (DoF) rendering and volume rendering. 
The opacity in volume rendering aligns with the layer visibility in DoF rendering.
To execute the blurring, we modify the layered blur kernel to the ray-based kernel and employ an optimized sparse kernel to gather the input rays efficiently and render the optimized rays with our layered DoF volume rendering. 
We synthesize a dataset with defocused dynamic scenes for our task, and extensive experiments on our dataset show that our method outperforms existing approaches in synthesizing all-in-focus novel views from defocus blur while maintaining spatial-temporal consistency in the scene.
% We synthesize a dataset with defocused dynamic scenes for our task, and extensive experiments on our dataset show that our method outperforms existing approaches in recovering a sharp representation and synthesizing all-in-focus novel views from defocus blur while maintaining spatial-temporal consistency in the scene.

\keywords{Dynamic Novel View Synthesis \and Neural Radiance Field \and Depth-of-Field}
\end{abstract}

%% file: sec/1_intro.tex
\section{Introduction}
\begin{figure}[t]
    \centering
    \includegraphics[width=0.9\linewidth]{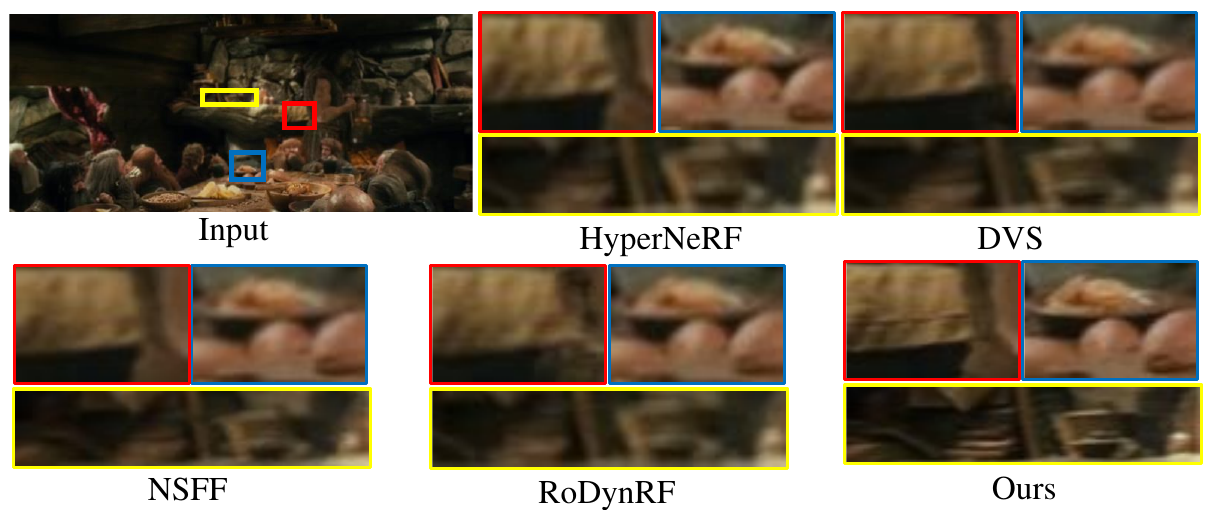}
    \caption{Given a monocular video captured with defocus blur, existing dynamic NeRF approaches fail to recover high-quality details and tend to produce blurry views, and our method \ourmethod synthesizes sharp novel views.}
    \label{fig:fig1}
    % \vspace{-10pt}
\end{figure}
\label{sec:intro}
Capturing videos from cameras or smartphones has become a new norm for daily life. However, videos are typically recorded from a monocular camera, restricting the captured scene to fixed viewpoints. To depict the dynamic scene from flexible viewpoints, dynamic view synthesis\cite{pumarola2021d,li2021neural,li2023dynibar} is proposed to facilitate the generation of photorealistic novel views from arbitrary camera angles and perspectives, thus allowing for free viewpoints.
The ability to capture a 3D dynamic scene in a monocular setting enhances the importance of dynamic novel view synthesis in video-related tasks such as video stabilization, augmented reality, and view interpolation. 
% Novel view synthesis\cite{debevec1996modeling,gortler1996lumigraph} finds widespread application in fields like augmented reality and virtual reality. 
% Specifically, dynamic novel view synthesis plays a crucial role in video-related tasks such as video stabilization and video interpolation. 
To interpolate new views from existing sparse input views, recent works use neural networks to train a neural radiance field (NeRF)~\cite{mildenhall2020nerf}. The network is a multilayer perceptron (MLP) that learns to map spatial coordinates and view directions to volume densities and view-dependent RGB colors. 

Recent dynamic NeRF methods~\cite{park2021hypernerf,li2021neural,li2023dynibar} have made progress by establishing spatial-temporal consistency to achieve space-time novel view synthesis. 
However, these methods are designed under the assumption that the model is trained on all-in-focus image sequences, where all objects within the scene remain focused and sharp during the entire filming.
Defocus blur has not been addressed by current dynamic NeRF models, and the blur can result in performance degradation for the reconstructed dynamic NeRF, as it poses challenges for modeling dynamic object motions due to the lack of sharp details and the inability to model temporal consistency.
As shown in Fig.~\ref{fig:fig1}, current dynamic NeRF methods perform poorly when applied to monocular videos affected by defocus blur, resulting in an inability to restore sharp contents.
Maintaining all objects in the scene in focus may offer a solution, however, maintaining every frame in focus is challenging when capturing monocular videos. 
Defocus blur occurs when the camera's aperture is wide, and the objects are not positioned at the focal distance. 
% Defocus blur occurs when the camera's aperture is wide, and the objects are not positioned at the focal distance which is the disparity of the focused object. 
Therefore, defocus blur tends to manifest in videos due to the scene comprising multiple objects with significant depth variations coupled with the common equipment of large aperture for more natural light and enhanced imaging quality. Furthermore, the focal distance may constantly fluctuate in frames if the photographer lacks expertise. 
% The blur can result in performance degradation for the reconstructed dynamic NeRF. 
Considering that defocus blur is almost unavoidable during video capture, recovering a sharp dynamic NeRF is a valuable problem yet to be explored.

Several works have been proposed to address defocus blur from static scenes. Physical scene prior and kernel estimation~\cite{ma2022deblur,Lee_2023_CVPR} are introduced to mitigate blurring, and an explicit scatter-gather technique~\cite{dofnerf} is proposed to better model the defocus blur. However, these methods are specifically tailored for static multi-view inputs and are not capable of representing defocused dynamic scenes. 
% In addition, current methods perform the blur in a post-process manner without fusing the defocus blur model in NeRF rendering.
Previous methods apply blurring after the last step of NeRF: volume rendering. The blurring is before optimization but after the typical rendering.

% For static NeRF, there have been several works proposed to synthesize high-quality novel views from abnormal inputs, \eg the absent poses~\cite{wang2021nerf,meng2021gnerf,jeong2021self}, illumination variation~\cite{Martin-Brualla_2021_CVPR}, exposure noise~\cite{huang2021hdr}, and aliasing~\cite{barron2021mip}. 
% To address defocused static scenes, physical scenes prior and kernel estimation~\cite{ma2022deblur,Lee_2023_CVPR} are introduced to mitigate blurring, and an explicit scatter-gather technique~\cite{dofnerf} is proposed to better model the defocus blur. However, these methods are specifically tailored for static NeRF and are not capable of representing defocused dynamic scenes. 
% Existing dynamic NeRF methods have not taken defocus blur into account. Defocus blur is harmful to modeling dynamic object motions because it interferes with the cross-time temporal consistency. 
% Existing dynamic NeRF methods have not yet addressed defocus blur, which poses challenges for modeling dynamic object motions due to the lack of sharp details and the inability to model temporal consistency.
% As shown in Fig.~\ref{fig:fig1}, current dynamic NeRF methods perform poorly when applied to monocular videos affected by defocus blur, resulting in an inability to restore sharp contents.

To tackle defocus blur in monocular dynamic videos, we propose \ourmethod, a dynamic framework designed to incorporate the defocus blur model into dynamic NeRF training, enabling the recovery of sharp novel views from defocused monocular videos. 
% To tackle this issue, we propose \ourmethod, a dynamic framework designed to incorporate the defocus blur model into dynamic NeRF training, enabling the recovery of sharp novel views from defocused monocular videos. 
Our approach involves modeling defocus blur by estimating blur kernels and introducing layered Depth-of-Field (DoF) volume rendering to the input rays. We connect the classical volume rendering with DoF rendering, seamlessly integrating the blurring process into the NeRF pipeline. 
We modify the layer-based kernel in DoF rendering to the ray-based kernel to fit into volume rendering in NeRF. To decrease computational costs,  
we transition from using a traditional disk blur kernel to employing a sparse one~\cite{ma2022deblur} and optimize the rays originating from these sparse points. We use an MLP to predict both the 3D positions of the kernel points and their corresponding weights.
Subsequently, the optimized rays and the estimated kernels serve as inputs for the aforementioned layered DoF volume rendering process. It is worth noting that our \ourmethod enables the dynamic NeRF model to learn a sharp scene representation from defocused input views, leading to sharp novel views.

To improve the quality of reconstructed static regions, we utilize the blending technique~\cite{li2021neural} to independently predict the static and dynamic scenes. The blended color is then rendered with our layered DoF volume rendering pipeline.
Furthermore, we adopt a cross-time rendering approach to represent temporal consistency in dynamic scenes. This approach models scene correlation across multiple input views and utilizes deformed rays for layered DoF volume rendering. Integrating scene information from other frames into the target frame enhances the representation of dynamic scenes. 

To address the absence of existing dynamic NeRF defocus deblurring datasets, 
we collect and synthesize defocused video sequences from a stereo dataset~\cite{wang2023neural} and conduct an extensive analysis to demonstrate \ourmethod can effectively address defocus blur inputs and render high-quality sharp novel views.
% we collect both synthetic and real-world datasets and conduct an extensive analysis to demonstrate \ourmethod can effectively address defocus blur inputs and render high-quality sharp novel views.
We compare with current dynamic NeRF methods as well as the combination of single-image deblurring and dynamic NeRF. Results show that \ourmethod outperforms existing methods.
Additionally, we conduct ablation studies to validate the effectiveness of each component. 
In summary, our main contributions are as follows.
\begin{itemize}
    \item[$\bullet$] We present the first dynamic NeRF model \ourmethod designed to recover sharp novel views from a defocused monocular video.
    \item[$\bullet$] We introduce layered DoF volume rendering in dynamic NeRF to model the defocus blur and restore a sharp representation.
    \item[$\bullet$] We transform the layered blur kernel to the ray-based kernel to fit into NeRF training and incorporate the optimized sparse kernel with the layered DoF volume rendering, enabling an efficient blurring process.
    % \item[$\bullet$] We also collect a new dataset for defocus deblurring of dynamic NeRF
\end{itemize}

%% file: sec/2_related.tex
\section{Related Work}
% \subsection{Depth-of-Field Rendering}
\noindent\textbf{Depth-of-Field Rendering.}
Depth-of-field (DoF) rendering, also known as bokeh rendering, generates realistic defocus blur
in out-of-focus areas to enhance the visual prominence of the focused subject within an image. Physically based methods~\cite{lee2010real,abadie2018advances} rely on 3D scene information and they are time-consuming. 
Recent methods use neural networks~\cite{ignatov2019aim,ignatov2020aim} for end-to-end training. DeepFocus~\cite{xiao2018deepfocus} entails an exact depth map to render realistic bokeh. However, the exact depth map is hard to acquire. Therefore, some methods~\cite{wang2018deeplens,wadhwa2018synthetic} instead predict depth for DoF rendering.  
To automate the rendering process, bokeh is directly rendered~\cite{ignatov2020rendering,luo2023defocus} from an all-in-focus input without depth maps and controlling parameters. Synthesized data~\cite{peng2022bokehme} from ray tracing is introduced for training to combine classical and neural rendering. 
To generate partial occlusion, some methods~\cite{peng2022mpib,sheng2023dr} decompose the scene into multiple layers and separately blur each layer before compositing them, inspired by layered rendering~\cite{busam2019sterefo,zhang2019synthetic}.
% To generate partial occlusion, MPIB~\cite{peng2022mpib} decomposes the scene into multiple layers and separately blurs each layer before compositing them, inspired by layered bokeh rendering~\cite{busam2019sterefo,zhang2019synthetic}.
In this work, we show the connection between DoF rendering and NeRF rendering, integrating the layered blurring kernels with volume rendering.

% \subsection{Defocus Deblurring}
\noindent\textbf{Image Defocus Deblurring.}
To recover a sharp image from defocus blur, traditional methods conduct a two-stage pipeline: (1) estimate a defocus map~\cite{shi2015just,park2017unified} from image priors, (2) the defocus map is used as guidance to deblur the image from non-blind deconvolution~\cite{levin2007image,dong2021dwdn}. 
With the current advances in neural networks, recent methods train end-to-end on datasets~\cite{abuolaim2020defocus}. Modifications on convolutions~\cite{lee2021iterative,son2021single} are proposed to improve the deblurring performance. The Transformer architecture and a novel multi-scale feature extraction module~\cite{zamir2022restormer} are introduced for high-resolution restoration. Additional inputs, such as dual pixel~\cite{abuolaim2020defocus}, light field~\cite{ruan2022learning}, video sequences~\cite{abuolaim2021learning}, and depth images~\cite{pan2021dual} are utilized to guide defocus deblurring. 
Current methods do not take multi-view 3D geometry and dynamic scenes into account. We aim to explore defocus deblurring in dynamic NeRF synthesis, and our method integrates the DoF rendering into dynamic NeRF to synthesize defocus blur, which in turn facilitates deblurring.

\noindent\textbf{Neural Radiance Field.}
Neural radiance field (NeRF) is an implicit representation~\cite{NEURIPS2019_SRN, mildenhall2019local, sitzmann2019deepvoxels, 2019Neural} for photo-realistic novel view synthesis, which aims to generate new camera perspectives given a specified set of input viewpoints. 
Different from explicit representations~\cite{jimenez2016unsupervised,fan2017point}, NeRF~\cite{mildenhall2020nerf} learns an implicit continuous function to model the complex geometry and appearance of a scene.
Subsequent studies aim to synthesize high-quality novel views from abnormal inputs. 
% Several works~\cite{wang2021nerf,meng2021gnerf,jeong2021self} address scenarios with no camera poses. 
To recover a sharp NeRF from defocus-blurred inputs, Deblur-NeRF~\cite{ma2022deblur} proposes a deformable sparse kernel to model defocus blur, and DP-NeRF~\cite{Lee_2023_CVPR} adopts physical scene priors to mitigate blurring. DoF-NeRF~\cite{dofnerf} introduces the Concentrate-and-Scatter technique to explicitly enable controllable DoF effects. However, all these methods render DoF in a post-process manner and do not consider the volume sampling process. LensNeRF\cite{kim2024lensnerf} restores a sharp NeRF from defocused static scene. The technique is valid for static scenes but does not model the temporal photometric consistency between video frames. Current methods are only effective in static scenes, where the objects are not moving. Dynamic NeRF synthesis with defocus blur remains unexplored. 
% We discuss more existing works related to the DoF task in supplementary. 

% \subsection{Dynamic Neural Radiance Field}
\noindent\textbf{Dynamic Neural Radiance Field.} 
Recent works extend NeRF from static scenes to dynamic scenes~\cite{xian2021space, li2023dynibar}. Given a monocular video, some approaches apply a deformation field on a canonical representation~\cite{pumarola2021d, tretschk2021non, park2021nerfies, park2021hypernerf, wang2023flow} for space-time novel view synthesis. Scene flow also works as a constraint to establish temporal consistency and model object motions in dynamic scenes~\cite{li2021neural, gao2021dynamic, wang2021neural, liu2023robust}. 
These methods require sharp inputs and fail when the defocus blur is in the input monocular video. Compared with existing methods, we synthesize novel views given the defocus blur. Furthermore, we propose to tackle the blur by introducing the layered DoF volume rendering in dynamic NeRF rendering.

%% file: sec/3_Method.tex
\section{Method}
\label{sec:method}

% 方法图

We propose \ourmethod, a dynamic radiance field pipeline to recover sharp novel views from a defocused monocular video. 
% As shown in Fig.~\ref{fig:pipeline}, we introduce layered DoF volume rendering, which incorporates layered DoF rendering into the volume rendering in dynamic NeRF. 
In the following, we first briefly revisit the concept of NeRF and the principle of DoF rendering. In section~\ref{sec:method_3}, we analyze the connection and the distinction between DoF rendering and volume rendering in NeRF training, proposing layered DoF volume rendering with optimized ray-based blur kernels. Then we show how to represent dynamic scenes in a space-time manner, blend dynamic and static scenes with layered DoF volume rendering, and ensure temporal consistency within the scene in Section~\ref{sec:method_4}. Finally, we provide the training details in Section~\ref{sec:method_5}.
% In the following, we first briefly revisit the concept of NeRF in section~\ref{sec:method_1} and the principle of DoF rendering in Section~\ref{sec:method_2}. Then in section~\ref{sec:method_3}, we analyse the connection between DoF rendering and volume rendering in NeRF training, proposing layered DoF volume rendering with optimized blur kernels. Finally, we show how to represent dynamic scenes in a space-time manner, blend dynamic and static scenes with layered DoF volume rendering, and ensure temporal consistency within the scene in Section~\ref{sec:method_4}. The training details are in Section~\ref{sec:method_5}.

\subsubsection{Preliminary: Neural Radiance Field}
\label{sec:method_1}
Our method follows the principle of Neural Radiance Field (NeRF), which represents a static $3$D scene as an implicit continuous function $F_{\Theta}$. This function is parameterized by a multilayer perceptron (MLP) and it takes a spatial point location $\vect{x} \in \mathbb{R}^3$ and viewing direction $\vect{d} \in \mathbb{R}^3$ as inputs. $F_{\Theta}$ maps the inputs to color $\vect{c}$ and volume density $\sigma$:
\begin{equation}
    (\vect{c}, \sigma) = F_{\Theta}(\vect{x}, \vect{d})\,.
\end{equation}

To acquire the RGB value of a pixel coordinate $p$, 
NeRF first specifies a ray $\vect{r}_p(t)=\vect{o}+t\vect{d}_p$ emitted from camera projection center $\vect{o}$ along the viewing direction $\vect{d}_p$. Then followed by the classical volume rendering technique~\cite{kajiya1984ray}, the color of $\vect{r}_p(t)$ can be represented as an integral of all colors along the ray, computed as: 
% It can also be approximated as the sum of weighted radiance at $k$ points $\left \{ t_i \right \}_{i=1}^k$ on ray $\vect{r}(t)$:
% on the image plane, 
% its color is computed as: 
% \begin{equation}
% \label{eq:volume rendering}
%     \mathbf{C}(\mathbf{r}) = \int_{i_n}^{i_f} T(i) \thinspace \sigma (\mathbf{r}(i)) \thinspace \mathbf{c}(\mathbf{r}(t_i), \mathbf{d}) di\,,
% \end{equation}
\begin{equation}
    \begin{aligned}
        \hat{C}(\vect{r}_p)=\sum_{i=1}^{k}T_i(1-\exp(-\sigma\delta_i))\vect{c}(\vect{r}_p(t_i),\vect{d}_p)\,, 
    \end{aligned}\label{eq:volume}
\end{equation}
where
\begin{equation}\label{eq:ti}
    T_i=\exp(-\sum_{j=1}^{i-1}\sigma \delta_j)\, 
\end{equation}
is the accumulated transmittance along ray $\vect{r}_p$, and $\delta_i=t_{i+1}-t_i$ is the distance between the two sampled points. The vanilla NeRF assumes a pinhole camera model, where the color of a pixel $\hat{C}(\vect{r}_p)$ is calculated by points on a single ray $\vect r_p(t)$, so it fails to model the defocus blur.

\subsubsection{Preliminary: Defocus Blur and DoF Rendering}
\label{sec:method_2}
% dof and inter-layer occlusion图
\begin{figure}
    \centering
    \includegraphics[width=0.88\linewidth]{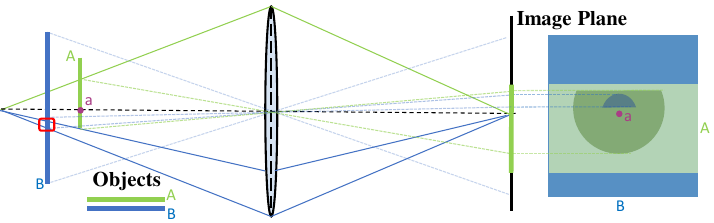}
    \caption{\textbf{The defocus blur formation.} 
    % We model the rendered defocus blur of the center purple pixel that gathers its neighboring pixels on the green foreground object and those on the blue background plane.
    We model the defocus blur of the center purple pixel. The blur and green objects are from different layers.
    We define the light path as the left-right view and the image plane shows the front-back view of the defocus blur.
    The CoC of the purple point gathers its neighboring pixels on the green foreground and those on the blue background.
    This defocus modeling indicates that the object originally occluded (red square from the blue object) by the green object under the pinhole camera view can contribute to the rendered purple point by the green and blue semi-circles. 
    % Therefore, we need to apply layer visibility $W_i$ in Eq.\ref{eq:bokeh_2} to correctly model the ray-gathering and establish the correct blending of layers. 
    In Eq.\ref{eq:bokeh}-\ref{eq:volume_2} we model the defocus blur from layer visibility $W_i$, then use the link between visibility and opacity to integrate DoF and volume rendering in Eq.\ref{eq:bokeh_volume}.
    The visibility is learned from the layered DoF volume rendering pipeline.
    % This is termed inter-layer occlusion, and we use layered bokeh rendering in Eq.\ref{eq:bokeh} to correctly model the occlusion and establish the correct blending of layers.
    }
    \label{fig:dof}
\end{figure}
% \begin{figure}
%     \centering
%     \includegraphics[width=1.0\linewidth]{images/kernel.pdf}
%     \caption{\textbf{Principle of Eq.\ref{eq:bokeh}.} 
%     }
%     \label{fig:kernel}
% \end{figure}
To model the defocus blur and render DoF, first we introduce bokeh rendering. Bokeh is the aesthetic quality of the blur in the out-of-focus region, caused by Circle of Confusion (CoC). Previous works~\cite{wadhwa2018synthetic,peng2022bokehme} have shown the CoC formation process. The radius $\gamma$ of each pixel is
\begin{equation}\label{eq:radius}
    \gamma=af\,\Big\lvert \frac{1}{z}-\frac{1}{z_f} \Big\rvert\,,
    % r=af\,\Big\lvert \frac{1}{z}-\frac{1}{z_f} \Big\rvert\,,
    % =A\,\lvert d-d_f \rvert\,,
\end{equation}
where $a$ is the camera aperture radius, $f$ is the focal length. 
% We replace them with $A=af$ to reflect the overall blur amount of the image. 
$z$ is the depth of the pixel and $z_f$ is the focal distance.
% We replace the depth $z$ with the disparity (inverse depth) $d$ for simplicity. 
Now that we have the blur radius of each pixel, we adopt layered bokeh rendering~\cite{peng2022mpib}. 
To blur the input image with kernels, classical bokeh rendering can be categorized into $gather$ and $scatter$ operations. We chose $gather$ because it fits the layered bokeh rendering pipeline and is compatible with the volume rendering in NeRF described in section~\ref{sec:method_3}. 
We define the layer as a set of fronto-parallel planes at fixed depths by discretization. Now we have the blur kernel $K(\gamma_i)$ for layer $i$, we apply an alpha composition for DoF rendering. The rendered DoF image $B$ is formulated as:
\begin{equation}\label{eq:bokeh}
    B=\frac{\psi(C_i)}{\psi(\mathbb I)}\,,
    % B=\frac{\sum^N_{i=1}\Big((C_iW_i*K(\gamma_i))\prod^N_{j=i+1}(1-W_j*K(\gamma_j))\Big)}{\sum^N_{i=1}\Big((W_i*K(\gamma_i))\prod^N_{j=i+1}(1-W_j*K(\gamma_j))\Big)}\,,
    % B=\frac{\sum^N_{i=1}\Big((C_iW_i*K(r_i))\prod^N_{j=i+1}(1-W_j*K(r_j))\Big)}{\sum^N_{i=1}\Big((W_i*K(r_i))\prod^N_{j=i+1}(1-W_j*K(r_j))\Big)}\,,
\end{equation}
where
\begin{equation}\label{eq:bokeh_2}
    \psi(X)=\sum^N_{i=1}\prod^N_{j=i+1}(1-W_j*K(\gamma_j))(XW_i*K(\gamma_i))\,.
    % B=\frac{\sum^N_{i=1}\Big((C_iW_i*K(\gamma_i))\prod^N_{j=i+1}(1-W_j*K(\gamma_j))\Big)}{\sum^N_{i=1}\Big((W_i*K(\gamma_i))\prod^N_{j=i+1}(1-W_j*K(\gamma_j))\Big)}\,,
    % B=\frac{\sum^N_{i=1}\Big((C_iW_i*K(r_i))\prod^N_{j=i+1}(1-W_j*K(r_j))\Big)}{\sum^N_{i=1}\Big((W_i*K(r_i))\prod^N_{j=i+1}(1-W_j*K(r_j))\Big)}\,,
\end{equation}
$\psi(X)$ is the compositing blurring function with the image $X$ as input, $*$ is the convolution operation, $N$ is the number of layers, $\mathbb I$ is an image with all values as $1$. 
Each layer $i$ encodes an RGB image $C_i$, with their corresponding visibility weight $W_i$, ensuring the correct blending of different layers.
% Each layer $i$ encodes an RGB image $C_i$, with their corresponding inter-layer occlusion weight~\cite{sheng2023dr} $W_i$, ensuring the correct blending of different layers.
We show the defocus blur formation and how rays gather between layers in Fig.~\ref{fig:dof}, demonstrating why we should apply the visibility term in DoF rendering. 
% We show the defocus blur formation and the inter-layer occlusion in Fig.~\ref{fig:dof}. 
This layered visibility-aware rendering gathers the rays with inter-layer weights for each layer, then blends the layers with the weights from back to front. 
% To handle the artifacts caused by layer discretization, we apply weight normalization in Eq.~\ref{eq:bokeh}. 
% This layered occlusion-aware bokeh rendering gathers the pixels with inter-layer occlusion for each layer, then blends the layers with inter-layer occlusion terms from back to front. To handle the artifacts caused by layer discretization, we apply weight normalization in Eq.~\ref{eq:bokeh}. 

\begin{figure}[t]
    \centering
    \includegraphics[width=0.95\linewidth]{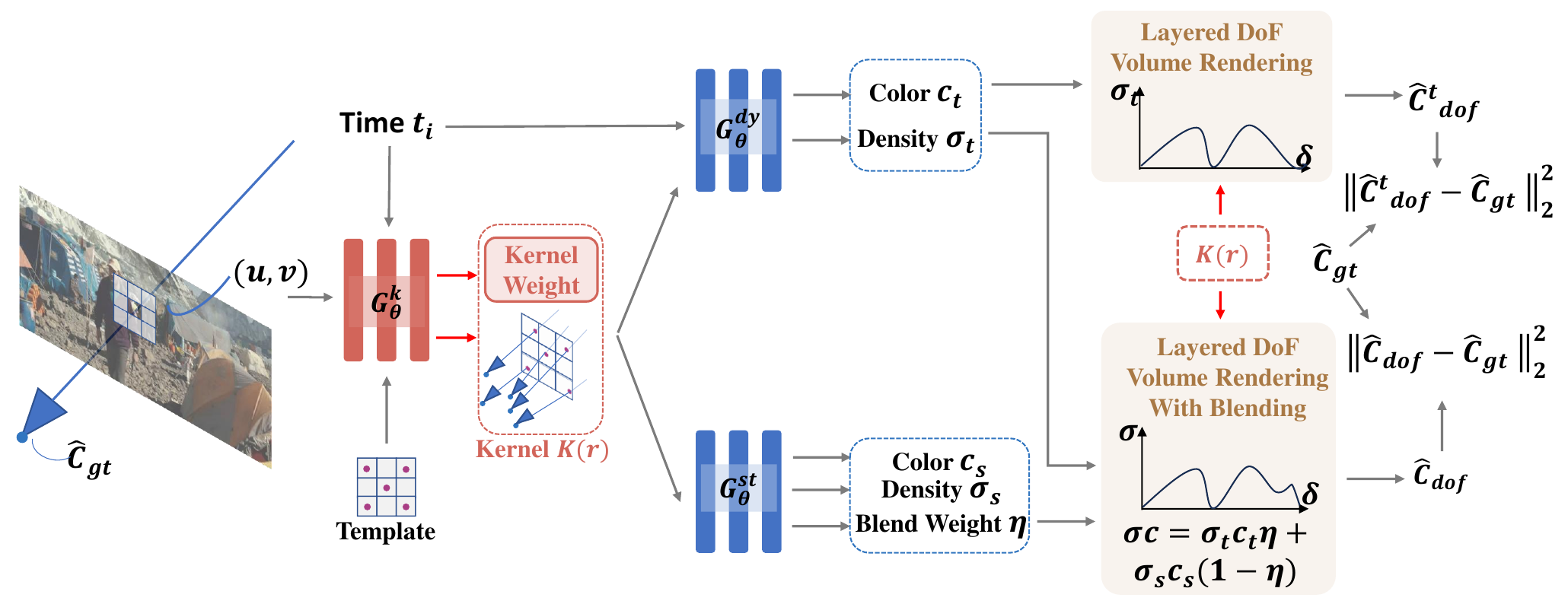}
    \caption{\textbf{Pipeline of our framework.} The framework takes a set of plane coordinates $(u,v)$, the time embedding $t_i$, and defines a kernel template as inputs, the outputs are the blur kernels $K(\vect{r_i})$ consisting of the sparse optimized rays with their corresponding weights. 
    The rays are then fed to the two MLPs $G_{\theta}^{\textrm{st}}$ and $G_{\theta}^{\textrm{dy}}$ to independently represent static and dynamic scenes. The final color is rendered by layered DoF volume rendering (Section~\ref{sec:method_3}), $\hat{C}_{dof}(\vect{r})$ is from Eq.\ref{eq:bokeh_volume} and $\hat{C}_{dof}^{t}(\vect{r})$ is from Eq.\ref{eq:blend}. The rendered defocused results (dynamic and blended) are supervised by the input defocused views. 
    For testing we directly render the rays without layered DoF volume rendering and the kernel.}
    % When testing, we directly render the rays without layered DoF volume rendering and the blur kernel.}
    \label{fig:pipeline}
\end{figure}

\subsection{Layered DoF Volume Rendering}
\label{sec:method_3}
Now that we have a basic understanding of NeRF and DoF rendering, we introduce our layered DoF volume rendering module. 
As shown in Fig.~\ref{fig:pipeline}, we introduce layered DoF volume rendering, incorporating DoF rendering into volume rendering in dynamic NeRF.
% When applying Eq.~\ref{eq:bokeh} and Eq.~\ref{eq:bokeh_2} to DoF rendering, we found that the blurring formation and volume rendering in Eq.~\ref{eq:volume} have connections. 
To seamlessly fuse Eq.~\ref{eq:bokeh} and Eq.~\ref{eq:bokeh_2} in NeRF training, we utilize the connections between the blur formation and volume rendering in Eq.~\ref{eq:volume}. 
The volume rendering assumes that, for each ray $\vect r(t_i)$ from a scene, the sampling points in the volume all have an alpha value 
% \begin{equation}
%     \begin{aligned}
%         \alpha_i=1-\exp(-\sigma\delta_i)\,,
%     \end{aligned}\label{eq:alpha}
% \end{equation}
$\alpha_i=1-\exp(-\sigma\delta_i)$, 
which indicates the opacity of the point. 
A larger alpha value shows greater absorption of light, making it more likely for the light to pass through that point. In contrast, a smaller alpha value means less absorption of light at that point, and the object at that point is more transparent.
In Eq.~\ref{eq:bokeh_2}, the term $W_i$ represents the visibility between the current layer $i$ and all the layers behind it. This visibility $W$ has the same physical meaning as the opacity value $\alpha$ in volume rendering. 
% In Eq.~\ref{eq:bokeh}, the inter-layer occlusion term $W_i$ represents the inter-layer visibility between the current layer and all the layers behind it. This inter-layer visibility $W$ has the same physical meaning as the opacity value $\alpha$ in volume rendering. 
Furthermore, in NeRF rendering, the volume of the scene is sampled in a discretized manner, which typically involves uniformly sampling discrete points along the ray path. This layer discretization is compatible with the one in DoF rendering.

Therefore we first transform Eq.~\ref{eq:bokeh} from the format of layered images into the format of individual layered pixels:
\begin{equation}\label{eq:volume_2}
\begin{aligned}
    % B(\vect{r})&=\frac{\sum^k_{i=1}\Big((\vect{c}_iw_i*K(\vect{r}))\prod^k_{j=i+1}(1-w_j*K(\vect{r}))\Big)}{\sum^k_{i=1}\Big(w_i*K(\vect{r}))\prod^k_{j=i+1}(1-w_j*K(\vect{r}))\Big)} \\
    B(\vect{r})=\frac{\sum^k_{i=1}\Big(\prod^k_{j=i+1}(1-w_j*K(\vect{r}))w_i\vect{c}_i*K(\vect{r})\Big)}{\sum^k_{i=1}\Big(\prod^k_{j=i+1}(1-w_j*K(\vect{r}))w_i*K(\vect{r})\Big)}
    \,,
\end{aligned} 
\end{equation}
where $\vect{c}_i$ and $w_i$ are the colors and layer visibilities of the pixels surrounding $\vect{r}$ within the blur kernel $K(\vect{r})$ .
Then we propose to integrate layered DoF volume rendering into dynamic NeRF training, the rendering can be formulated as:
% Therefore, we propose to integrate layered DoF volume rendering into dynamic NeRF training, the rendering can be formulated as:
\begin{equation}\label{eq:bokeh_volume}
\begin{aligned}
    \hat{C}_{dof}(\vect{r}_p)=\frac{\sum^k_{i=1}\Big((T_i*K(\vect{r}))(1-\exp(-\sigma\delta_i))\vect{c}(\vect{r}(t_i),\vect{d})*K(\vect{r}))\Big)}{\sum^k_{i=1}\Big((T_i*K(\vect{r}))(1-\exp(-\sigma\delta_i))*K(\vect{r}))\Big)} 
    \,,
    % \hat{C}_{dof}(\vect{r}_p)&=\frac{\sum^k_{i=1}\Big((\vect{c}_i\alpha_i*K(r_i))\prod^k_{j=i+1}(1-\alpha_j*K(r_j))\Big)}{\sum^k_{i=1}\Big(\alpha_i*K(r_i))\prod^k_{j=i+1}(1-\alpha_j*K(r_j))\Big)} \\
    % &=\frac{\sum^k_{i=1}\Big((T_i*K(\vect{r}))(1-\exp(-\sigma\delta_i))\vect{c}(\vect{r}(t_i),\vect{d})*K(\vect{r}))\Big)}{\sum^k_{i=1}\Big((T_i*K(\vect{r}))(1-\exp(-\sigma\delta_i))*K(\vect{r}))\Big)} 
    % \,,
    \end{aligned} 
\end{equation}
% where 
% \begin{equation}\label{eq:ti}
%    T_i^k=\exp(-\prod_{j=1}^{i-1}\sigma \delta_j)*K(\vect{r}))\, 
% \end{equation}
where the variable name is consistent as Eq.~\ref{eq:volume} and Eq.~\ref{eq:volume_2}, \eg $k$ is the number of sampling layers both in DoF rendering and volume rendering, $\vect{c}_i$ is the emitted color of layer $i$ blurred by $K(\vect{r})$. 
% where the variable name is consistent as Eq.~\ref{eq:volume} and Eq.~\ref{eq:volume_2}, \eg $k$ is the number of sampling layers both in bokeh rendering and volume rendering, $\vect{c}_i$ is the emitted color of layer $i$ blurred by $K(\vect{r})$ corresponding to $C_i$ in Eq.~\ref{eq:bokeh}. 

Although DoF rendering and NeRF volume rendering share similar traits, 
there is a difference in their implementations, as the DoF rendering is originally used on different layers of the entire image, and NeRF training is optimized by rays sampled across the near and far planes. 
To implement the DoF rendering in Eq.~\ref{eq:bokeh_volume}, we switch the layer-based kernels $K(\gamma_i)$ to ray-based kernels $K(\vect{r})$, ensuring that the blur kernels are still diverse based on different positions. $K(\gamma_i)$ is defined by the divided layers, and $K(\vect{r})$ is defined by the input rays.
Furthermore, to alleviate the high computational cost, we apply a sparse kernel following~\cite{ma2022deblur}, which replaces the dense kernel with a smaller number of sparse points. The convolution process with $K(\vect{r}_p)$ can be formulated as:
% Furthermore, to alleviate the high computational cost, we apply a sparse kernel following~\cite{ma2022deblur}, which replaces the dense $\gamma \times \gamma$ blur kernel with a smaller number of sparse points. The convolution process with $K(\vect{r}_p)$ can be formulated as:
% As for the blur kernel $K(\vect{r})$, to alleviate the high computational cost, we apply a sparse kernel following~\cite{ma2022deblur}, which approximates the dense $r \times r$ blur kernel with a smaller number of sparse points. The convolution process with $K(\vect{r}_p)$ can be formulated as:
\begin{equation}\label{eq:kernel}
    \vect{b}_p=\sum_{j \in S(p)}\vect{c}_j g_j \,,
\end{equation}
where $\vect{b}_p$ is the result of point $p$ after convolution, $S(p)$ represents the sparse points surrounding point $p$, ${g_j}$ and $\vect{c}_j$ are the corresponding weights and colors from the sparse points. We set $\sum_{j \in S(p)}g_j=1$ to ensure the brightness consistency. Though we fix the number of sparse points, we additionally apply ray deformation as ~\cite{ma2022deblur} to optimize the rays as the spatially varying real-world blur kernels. 
As shown in Fig.~\ref{fig:pipeline}, 
we jointly optimize the translation of the center ray origin of the blur kernel. For a plane coordinate at the center of kernel $(u,v)$, we use an MLP $G_{\theta}^{\textrm{k}}$ to predict the deformation of the kernel points as well as the kernel weights:
\begin{equation}\label{eq:kernel}
    (\Delta\vect{j}, g_j) = G_{\theta}^{\textrm{k}}((u,v), \vect{j}, t_l), \,,
    % (\Delta\vect{j},\Delta\vect{o}_j, g_j) = G_{\theta}^{\textrm{k}}(\vect{p}, \vect{j}', t_l), \,,
\end{equation}
where $t_l$ is the embedding of the input frame, 
% $\vect{o}$ is the ray origin,
$\vect{j}$ are the original rays from the kernel template, $g_j$ are the corresponding kernel weights, and the final optimized ray inputs are calculated as 
% $\vect{j}=\vect{j}'+\Delta\vect{j}$
$\vect{r}_j=\vect{j}+\Delta\vect{j}$. 
% $\vect{r}_j=\vect{j}'+\Delta\vect{j}=(\vect{o}_j+\Delta\vect{o}_j)+t\vect{d}_j$. 
% $\vect{j}=\vect{j}+\Delta\vect{j}$. 

\subsection{Dynamic Radiance Field}
\label{sec:method_4}
With the layered DoF volume rendering module in place, we turn to the dynamic scene representation.
Given a monocular video of a dynamic scene, we take the video frames and camera parameters as inputs. Following previous dynamic NeRF methods,
we represent the scene with an MLP $G_{\theta}^{\textrm{dy}}$, which maps the input spatial point location $\vect{x}$, viewing direction $\vect{d}$, and time $t$, to the corresponding color, volume density, scene flow, and disocclusion weight:
\begin{equation}
    \begin{aligned}
        (\vect{c}_t, \sigma_t, f_t,\mathcal{W}_t) = G_{\theta}^{\textrm{dy}}(\vect{x}, \vect{d}, t), \,, 
    \end{aligned} \label{eq:naive}
\end{equation}
where $G_{\theta}^\textrm{dy}$ is parameterized by $\theta$. The scene flow $f_t$ and the disocclusion weight $\mathcal{W}_t$ are used for the following cross-time rendering. Based on the representation, we next describe our twist on the dynamic radiance field.

\subsubsection{Blending Dynamic and Static Scene.}
We use two MLPs to model static and dynamic scenes as shown in Fig.\ref{fig:pipeline}. One MLP is designed for modeling moving objects by various loss terms (cross-time \etc), and the other learns the static scene, allowing for more stable rendered still objects. 
To improve the quality of time-invariant static regions, we apply a blending technique as previous dynamic methods~\cite{li2021neural, liu2023robust}, using an additional MLP to represent the static scene: 
\begin{equation}
    (\vect{c}, \sigma, \eta) = G_{\theta}^{\textrm{st}}(\vect{x}, \vect{d})\, ,
\end{equation}
where $\eta$ denotes the blending weight used to blend static and dynamic models. 
We denote the static color as $\vect{c}_s$, the dynamic color as $\vect{c}_t$, the static volume density and the dynamic volume density as $\sigma_s$ and $\sigma_t$.
As shown in Fig.~\ref{fig:pipeline}, the layered DoF volume rendering with the blending weight is defined as:
\begin{equation}
    % \begin{aligned}
    \hat{C}_{dof}(\vect{r})=\frac{\sum^k_{i=1}T_i^bvolume_{i}^{\epsilon}}{\sum^k_{i=1}T_i^bvolume_{i}^{\varepsilon}} 
    \,,
    \label{eq:blend}
    % \end{aligned}
\end{equation}
where 
$T_i^b=\exp(-\sum_{j=1}^{i-1}\sigma_s \sigma_t\delta_j)*K(\vect{r})$.
$volume_{i}^{\epsilon}$ and $volume_{i}^{\varepsilon}$ are rendered with $\eta(t)$ as weights:
\begin{equation}
    \begin{aligned}
    volume_{i}^{\epsilon}&=\eta(t)(1-\exp(-\sigma_s\delta_i))\vect{c_s}(\vect{r}(t),\vect{d})*K(\vect{r}) \\
    &+(1-\eta(t))(1-\exp(-\sigma_t\delta)\vect{c_t}(\vect{r}(t),\vect{d}) *K(\vect{r}) 
    \,,
    \label{eq:weight}
    \end{aligned}
\end{equation}
and
\begin{equation}
    \begin{aligned}
    volume_{i}^{\varepsilon}&=\eta(t)(1-\exp(-\sigma_s\delta_i))*K(\vect{r}) \\
    &+(1-\eta(t))(1-\exp(-\sigma_t\delta)*K(\vect{r})  
    \,.
    \label{eq:normal}
    \end{aligned}
\end{equation}

Finally, we apply the rendering loss for the blending results: 
\begin{equation}
    \begin{aligned}
        \mathcal{L}_{\textrm{color}}^{\textrm{b}} = \Vert \hat{C}_{dof}(\vect{r}) - \hat{C}_{\textrm{gt}}(\vect{r}) \Vert_2^2\,.
    \end{aligned}
\end{equation}
Apart from blending the static and dynamic scenes, we also constrain the rendered result of dynamic scene $\hat{C}^{t}_{dof}(\vect{r})$ by $\mathcal{L}_{\textrm{color}}^{\textrm{t}} = \Vert \hat{C}^{t}_{dof}(\vect{r}) - \hat{C}_{\textrm{gt}}(\vect{r}) \Vert_2^2$.

\subsubsection{Cross-time rendering.}
Suppose we only apply photometric consistency on individual time stamps like static NeRF. In that case, the model might fail to build temporal consistency in dynamic scenes, resulting in overfitting input views and failing to synthesize correct novel views. Therefore, temporal consistency between neighboring frames~\cite{gao2021dynamic, li2023dynibar} is crucial for training dynamic radiance fields. 
To establish temporal consistency between corresponding frames, we apply cross-time rendering under layered DoF volume rendering. For a timestamp $t_m$ from the $m$-th frame, we assign the three-dimensional scene flow of a spatial point $\vect{x}$ from $t_m$ to timestamp $t_n$ as $f_{\vect{x}}^{m}(n)$, 
% \begin{equation}
%     f_{\vect{p}}^{a}(j) = \mathcal{T}_{\mathbf{x}}^{t_a}(t_b) - \mathcal{T}_{\mathbf{x}}^{t_a}(t_a)\, ,
%     \label{eq:scene flow}
% \end{equation}
where $n$ denotes the neighbor frames of $m$. 
The corresponding point of $\vect{x}$ at $t_n$ can be computed as $\vect{x}^{m \rightarrow n} = \vect{x} + f_{\vect{x}}^{m}(n)$. This correspondence works on all the sampled points of a ray, therefore, 
we use the layered DoF volume rendering to acquire the color of the warped ray:
% we use the mentioned layered DoF volume rendering to acquire the color of the warped ray:
% Establishing such relationships for every point on a ray $\mathbf{r}$ can obtain a warped ray $\mathbf{r}_l^{i \rightarrow j}$ from $t_l^i$ to $t_l^j$. Subsequently, we utilize volume rendering to obtain the color value of the warped ray $\mathbf{C}_l^{j \rightarrow i}(\mathbf{r})$: 

% \begin{equation}
%     \label{eq:warped}
%     \hat{C}_{dof}^{b \rightarrow a}(\vect{r}) =\frac{\sum^k_{i=1}\Big((T_i^a*K(r))(1-\exp(-\sigma^b\delta_i))\vect{c}^{b}(\vect{r}(t_i)^{a \rightarrow b},\vect{d})*K(r))\Big)}{\sum^k_{i=1}\Big((T_i^a*K(r))(1-\exp(-\sigma^b\delta_i))*K(r))\Big)} 
%     \,,
% \end{equation}
\begin{equation}
    \label{eq:warped}
    \hat{C}_{dof}^{n \rightarrow m}(\vect{r}) =\frac{\sum^k_{i=1}\Big((T_i^n*K(r))(1-\exp(-\sigma_n\delta_i))\vect{c_t}(\vect{r}(t_i)^{m \rightarrow n},\vect{d})*K(r))\Big)}{\sum^k_{i=1}\Big((T_i^n*K(r))(1-\exp(-\sigma_n\delta_i))*K(r))\Big)} 
    \,,
\end{equation}
where $T_i^n=\exp(-\sum_{j=1}^{i-1}\sigma_n \delta_j)$, and $\vect{r}(t_i)^{m \rightarrow n}=\vect{r}(t_i)^{m}+f_{\vect{p}}^{m}(n)$. Then we compare the rendering loss between $\hat{C}_{dof}^{n \rightarrow m}(\vect{r})$ and $\hat{C}_{\textrm{gt}}^m(\vect{r})$: 

\begin{equation}
    \mathcal{L}_{\textrm{cross}} = \sum_{n \in \mathcal{N}(m)} \mathbf{W}^{n \rightarrow m}(\vect{r}) \Vert \hat{C}_{dof}^{n \rightarrow m}(\vect{r}) - \hat{C}_{\textrm{gt}}^m(\vect{r}) \Vert_2^2 \, ,
\end{equation}
where $n \in \mathcal{N}(m)$ are the neighboring frames of the $m$-th frame, 
$\mathbf{W}^{n \rightarrow m}(\vect{r})$ is calculated from volume rendering the weight $\mathcal{W}_t$ in Eq.~\ref{eq:naive}. 
% $\mathbf{W}^{n \rightarrow m}(\vect{r})$ is the motion disocclusion weight mentioned in Eq.~\ref{eq:naive}. 
The principle of disocclusion weight comes from the ambiguity of scene flow. Although scene flow is effective in establishing pixel-level alignment, it performs poorly when the pixels in one image are occluded. Therefore, we need a confidence map to demonstrate whether the pixel is occluded and the flow is correct. A possible solution is to use forward-backward consistency check~\cite{chen2016full} to alleviate the problem. 
Here we let the network learn the weight $\mathbf{W}$ in an unsupervised manner~\cite{li2021neural}.

\subsection{Optimization}
\label{sec:method_5}
Monocular reconstruction of complex dynamic scenes is highly ill-posed and prone to local minima during optimization~\cite{gao2021dynamic,li2021neural}. Therefore, we introduce data-driven priors $\mathcal{L}_{\textrm{data}}$ to supervise the scene geometry apart from the previous photometric constraint. 
Similar to previous works~\cite{li2021neural}, $\mathcal{L}_{\textrm{data}}$ consists of (1) scale-shift invariant monocular depth loss and (2) scene flow consistency and L$1$ regularization. 
We use RAFT~\cite{teed2020raft} and DPT~\cite{ranftl2021vision} to obtain ground truth optical flow and depth for the input images.

\subsubsection{Implementations.}
We set the max kernel size as $10$, which denotes the sparse kernel points are inquired under a radius of $10$. The number of kernel points is $5$. We employ the Adam optimizer~\cite{kingma2014adam} to jointly optimize the static and dynamic MLPs and the blur kernels. 
We use COLMAP~\cite{schonberger2016structure} to estimate the camera intrinsics and extrinsics. The learning rate is $5 \times 10^{-4}$. We train each scene for $250k$ iterations, which takes around two days on a single NVIDIA RTX $3090$ GPU. The rendering takes roughly $13$ seconds for each $940\times360$ frame.
% We use COLMAP~\cite{schonberger2016structure} to estimate the camera intrinsics and extrinsics, and these parameters are fixed during optimization. The learning rate is $5 \times 10^{-4}$. We train each scene for $250k$ iterations, which takes around two days on a single NVIDIA RTX $3090$ GPU. The rendering takes roughly $13$ seconds for each $940\times360$ frame.

%% file: sec/4_exp.tex
\section{Experiments}
% \subsection{Dataset Collection}
There are no available datasets for defocused inputs in dynamic NeRF, 
so we curated $8$ defocused dynamic scenes from an existing dataset VDW~\cite{wang2023neural} for evaluation. 
% so we have curated a collection of $12$ defocused dynamic scenes, $8$ from an existing Dataset VDW~\cite{wang2023neural} and $4$ real-world dataset for evaluation. 
VDW consists of sharp stereo image sequences and their corresponding disparity sequences. 
% For the synthetic dataset, VDW consists of sharp stereo image sequences and their corresponding disparity. 
We then use the bokeh rendering pipeline in BokehMe~\cite{peng2022bokehme} for defocus blur generation. Given the disparity sequences, the synthesized bokeh quality surpasses current classical rendering and neural rendering methods. 
We collected $8$ dynamic scenes of $1880\times720$ suitable for our task from this dataset, the image is resized to $940\times360$ for training. 
To closely resemble actual capturing scenarios, unlike static blur dataset~\cite{ma2022deblur} with a random focusing scheme, we gradually adjust the focal distance along the scene disparity, mimicking the typical focusing process observed in videos. The aperture is fixed for each scene.
We utilize the left-view defocused image sequences for training and the corresponding right-view sharp image sequences for evaluation. The $8$ scenes are named Camp, Shop, Car, Mountain, Dining1, Dining2, Dock, Gate.
% As for the real-world dataset, we use a DSLR camera to capture monocular videos with defocus blur, we resize the frames to $960\times540$.

\subsection{Baseline Comparisons}
The baselines include two types: 1) state-of-the-art Dynamic NeRF, \eg scene flow methods NSFF~\cite{li2021neural} and DVS~\cite{gao2021dynamic}, deformation-based method HyperNeRF~\cite{park2021hypernerf}, robust dynamic method RoDynRF~\cite{liu2023robust}
and 2) an image-space baseline that uses single-view image deblurring~\cite{son2021single} then trains the dynamic NeRF with the deblurred inputs. As all the current dynamic methods are not designed for defocus blur, we choose the robust RoDynRF and the flow-based DVS to train with~\cite{son2021single} for baseline comparison.
We use the pre-trained deblurring model for fairness.

\subsection{Qualitative Results}
\begin{figure}[!h]
    \centering    \includegraphics[width=0.88\linewidth]{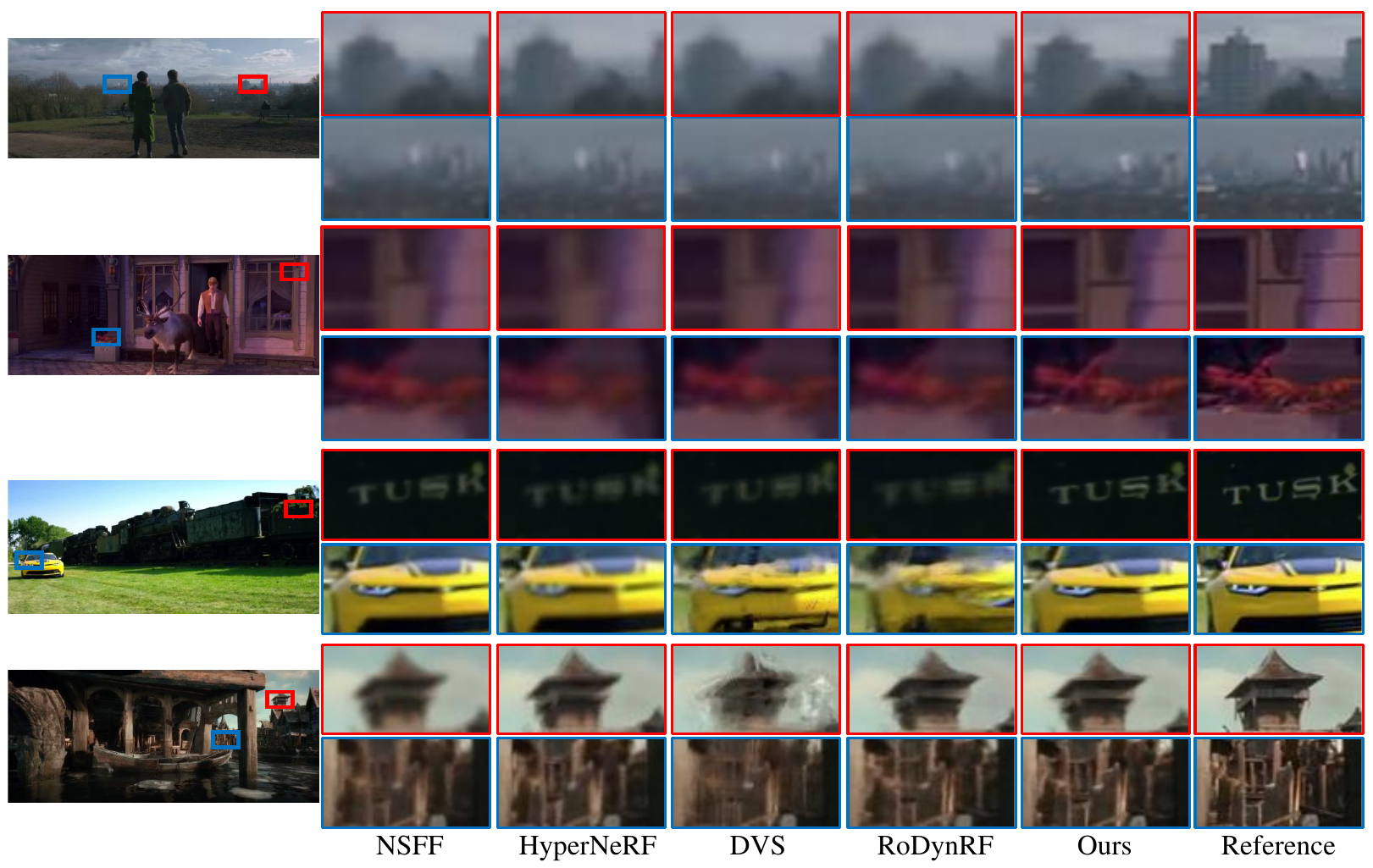}
    \caption{\textbf{The qualitative results with all dynamic NeRF baselines.} Compared with existing dynamic NeRF methods, our method generates sharper novel views that are more faithful and have more details. The scenes are Mountain, Shop, Car, Dock.
    }
    \label{fig:baseline_dynamic}
    % \vspace{-10pt}
\end{figure}

\begin{figure}[!h]
    \centering
    \includegraphics[width=0.88\linewidth]{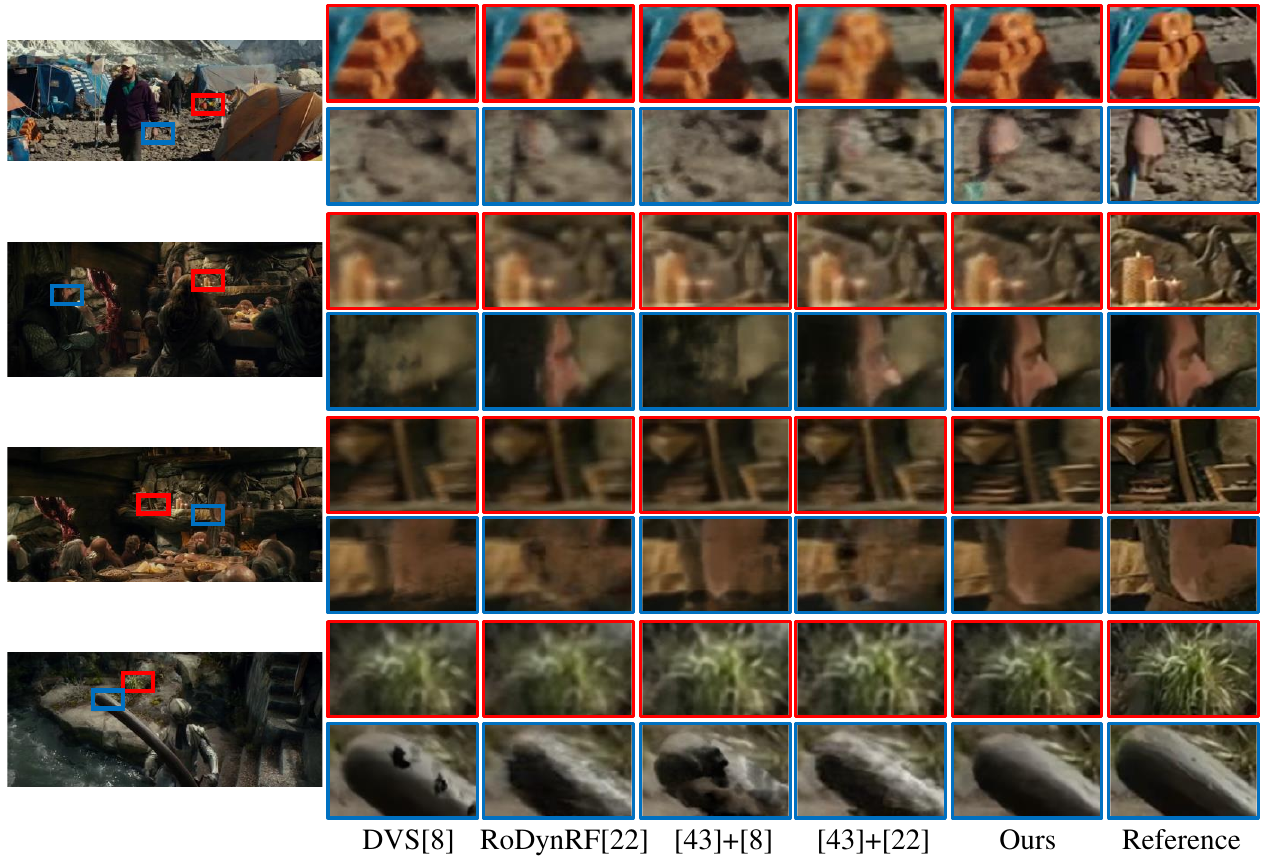}
    \caption{\textbf{The qualitative results with dynamic NeRF and their corresponding 2D image deblurring baselines.} Although 2D image deblurring helps to alleviate the blur for dynamic NeRF in novel views (red box), our method is more stable and generates more reliable sharp details. The scenes are Camp, Dining1, Dining2, Gate.
    % \caption{\textbf{The qualitative results with dynamic NeRF and their corresponding 2D image deblurring baselines.} Although 2D image deblurring helps to alleviate the blur for their counterpart NeRF in novel views (red box), our method is more stable and generates more reliable sharp details. The scenes are Camp, Dining1, Dining2, Gate.
    }
    \label{fig:baseline_deblur}
    % \vspace{-10pt}
\end{figure}

We show qualitative comparisons in Fig.~\ref{fig:baseline_dynamic} and Fig.~\ref{fig:baseline_deblur}. 
In Fig.~\ref{fig:baseline_dynamic}, we compare our method with existing dynamic NeRF methods, and in Fig.~\ref{fig:baseline_deblur}, we compare our approach with the chosen dynamic NeRF models and their counterparts with 2D image-deblurring for preprocessing. We present the results on all scenes in these two figures to show our method's effectiveness, $4$ scenes for each.

One can observe: 
1) current dynamic NeRF methods can not recover sharp details from the defocus blur; 2) the single-view 2D deblurring preprocessing~\cite{son2021single} can somewhat recover sharp details in some regions, however, each view with the deblurred content lacks consistency across views, and the performance is still less stable and reliable than our method;
3) our method \ourmethod outperforms these baselines in handling defocus blur and generating sharp novel views.  please refer to the supplementary material for more results.

\begin{table}[t]
\caption{\textbf{Quantitative comparisons against all baselines.} The best performance is \textbf{boldfaced}, and the second is \underline{underlined}.}
    \centering
    \setlength{\tabcolsep}{1.1pt}
    \renewcommand\arraystretch{1.2}
    \subfloat{
    \begin{tabular}{@{}lccc@{}}
        \toprule
        Methods & \scriptsize PSNR$\uparrow$ & \scriptsize SSIM$\uparrow$ & \scriptsize LPIPS$\downarrow$ \\
        \midrule
        HyperNeRF\cite{park2021hypernerf} & 26.96 & 0.780 & \underline{0.208} \\
        RoDynRF\cite{liu2023robust} & 26.18 & 0.770 & 0.227 \\
        NSFF\cite{li2021neural} & \underline{27.01} & \underline{0.803} &  0.209\\
        DVS\cite{gao2021dynamic} & 25.43 & 0.764 & 0.242 \\
        \midrule
        \ourmethod (Ours) & \textbf{27.30} & \textbf{0.816} & \textbf{0.130} \\
        \bottomrule
        % \multicolumn{4}{c}{(a)}
    \end{tabular}}\hskip1.2pt
    \subfloat{
    \begin{tabular}{@{}lccc@{}}
        \toprule
        Methods & \scriptsize PSNR$\uparrow$ & \scriptsize SSIM$\uparrow$ & \scriptsize LPIPS$\downarrow$ \\
        \midrule
        RoDynRF\cite{liu2023robust} & \underline{26.18} & 0.770 & 0.227 \\
        DVS\cite{gao2021dynamic} & 25.43 & 0.764 & 0.242 \\
        \cite{son2021single} + \cite{liu2023robust} & 25.79 & \underline{0.776} & \underline{0.196}\\
        \cite{son2021single} + \cite{gao2021dynamic} & 24.52 & 0.757 & 0.208 \\
        \midrule
        \ourmethod (Ours) & \textbf{27.30} & \textbf{0.816} & \textbf{0.130} \\
        \bottomrule
        % \multicolumn{4}{c}{(b)}
    \end{tabular}}
    \label{tab:quantitative}
\end{table}

\subsection{Quantitative Results}
We compare \ourmethod with the $6$ baseline methods quantitatively on our synthesized dataset.
We use PSNR, SSIM, and LPIPS~\cite{zhang2018unreasonable} as metrics for evaluation. 
% To evaluate the performance of these methods, we use PSNR, SSIM, and LPIPS~\cite{zhang2018unreasonable} as metrics. 

As shown in Table~\ref{tab:quantitative}, \ourmethod outperforms other state-of-the-art baselines. As mentioned above, to demonstrate the superiority of our method in modeling defocus blur in $3$D space, we choose two dynamic NeRF baselines RoDynRF and DVS, and train these dynamic NeRF models while preprocessing the input blurry images with $2$D defocus deblurring for comparison.
This comparison highlights that, although the $2$D deblurring method alleviates the defocus blur on each input view, this method struggles to maintain the consistency of scene information and tends to be unstable across all scenes. When dynamic NeRFs are trained on images deblurred with $2$D methods, it results in undesirable artifacts in scene geometry and unsatisfying performance overall. In contrast, our method models defocus blur in 3D space across all input views, ensuring spatial-temporal consistency of scene geometry. We show the average results on all scenes, please 
refer to the supplementary material for separate results on all scenes.

\subsection{Ablation Study}
\begin{figure}[t]
    \centering
    \includegraphics[width=0.9\linewidth]{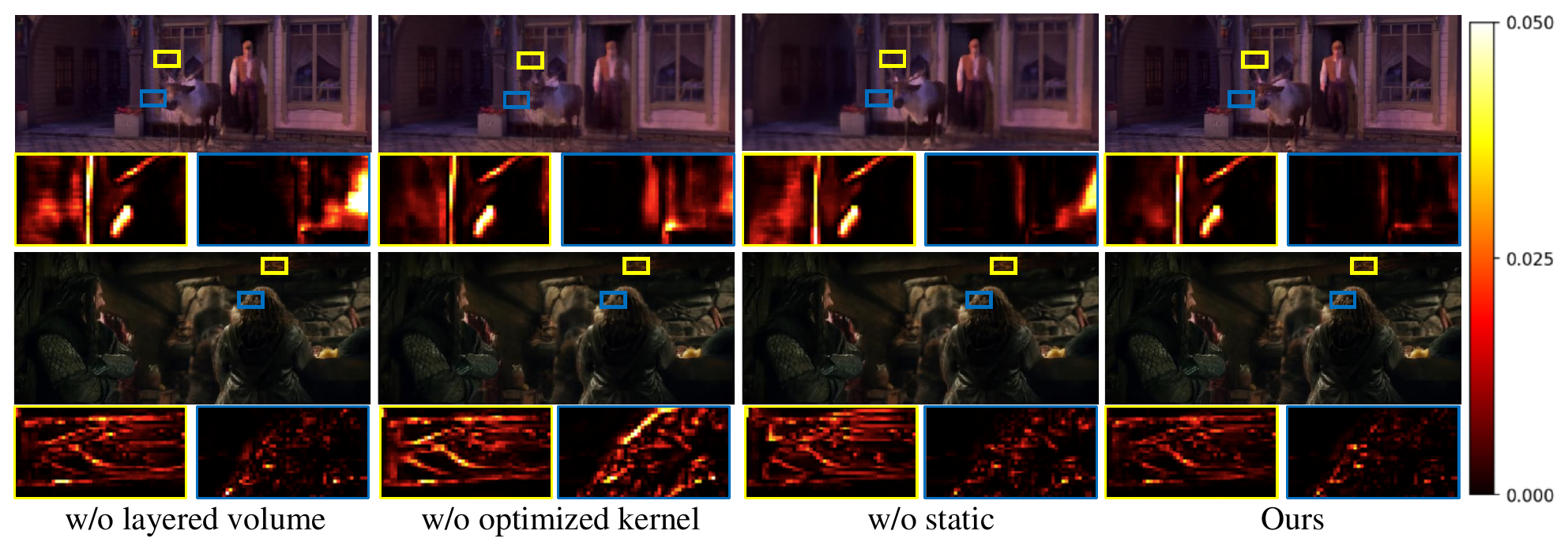}
    \caption{\textbf{The visualizations of the ablation study.} The corresponding error map is visualized at the bottom, where darker regions indicate smaller errors. 
    % We define the error range from $0$ to $0.05$. Our full model has the smallest error overall. 
    }
    \label{fig:ablation}
\end{figure}

The main idea of \ourmethod is to model the defocus blur by optimizing sparse blur kernels and incorporating the blur model into NeRF training. 
% The main idea of \ourmethod is to model the defocus blur to recover sharp details, and our method tackles the challenge by optimizing sparse blur kernels and incorporating the defocus modeling into NeRF training. 
Therefore we conduct ablation studies on the layered DoF volume rendering module and the optimized kernels. Furthermore, we evaluate the effects of the static-dynamic blending scheme.
% We visualize the results in Fig.~\ref{fig:ablation}. 
As shown in Fig.~\ref{fig:ablation} and Table~\ref{tab:ablation study}, our framework works best when all the components are applied. 
We remove (1) the layered volume rendering module, and we use the kernel weight to directly gather the final rendered color as post-processing; 
% We remove (1) the layered volume rendering module, and we use the kernel weight to directly gather the final rendered color as~\cite{ma2022deblur}; 
(2) the optimized kernel, and we apply the full regular kernel and learn bokeh parameters; 
% (2) the optimized kernel, and we apply the full regular kernel from learned bokeh parameters as~\cite{dofnerf}; 
(3) static representation.
We predict motion masks~\cite{li2021neural} to calculate the results on dynamic regions.
% We remove For the model without the layered volume rendering module, we use the kernel weight to directly gather the final rendered color as~\cite{ma2022deblur}. The model without the optimized kernel applies the full regular kernel from learned bokeh parameters as~\cite{dofnerf}. 
% We show that \ourmethod performs better on the edges of objects from the visualizations. 
We show that (1) the layered DoF volume rendering helps to model a more accurate defocus blur by integrating the blur into volume rendering, facilitating the recovery of sharp novel views; (2) the optimized kernel along with the layered DoF volume rendering provides an efficient and effective blurring pipeline for training; (3) the reconstruction of the static scene improves the performance because some regions are static and an extra scene representation helps to stabilize training.
We choose two scenes for visualization and show average quantitative results on all scenes, please refer to the supplementary material for more results.
% \vspace{-10pt}
% We choose two scenes: Dining1 and Shop for visualization and quantitative results, please
% refer to the supplementary material for results on all scenes.

% \begin{table}[t]\footnotesize
\begin{table}[t]
\caption{\textbf{Ablation study.} The best performance is \textbf{boldfaced}, and the second is \underline{underlined}. The left part calculates the results in dynamic regions, and the right part shows the results on the whole image.}
    \centering
    \setlength{\tabcolsep}{1.1pt}
    \renewcommand\arraystretch{1.2}
    \subfloat{
    \begin{tabular}{@{}lccc@{}}
        \toprule
        \scriptsize Method (Dynamic) & \scriptsize PSNR$\uparrow$ & \scriptsize SSIM$\uparrow$ & \scriptsize LPIPS$\downarrow$ \\
        \midrule
        w/o cross-time& {19.46}& 0.607&0.351  \\
        w/o layered volume& {24.42}& \underline{0.738}&\underline{0.170}  \\
        w/o optimized kernel & \underline{24.54} & 0.735 & 0.246\\
         w/o static & 23.93 & 0.719 & 0.179 \\
        \midrule
         Full (Ours) & \textbf{24.60} & \textbf{0.747} & \textbf{0.162} \\
        \bottomrule
        % \multicolumn{4}{c}{(a)}
    \end{tabular}}\hskip1.2pt
    \subfloat{
    \begin{tabular}{@{}lccc@{}}
        \toprule
        \scriptsize Method (All) & \scriptsize PSNR$\uparrow$ & \scriptsize SSIM$\uparrow$ & \scriptsize LPIPS$\downarrow$ \\
        \midrule
         w/o cross-time & 22.61& 0.725& 0.232\\
         w/o layered volume & 27.11& \underline{0.811} & 0.211\\
        w/o optimized kernel &\underline{27.25}& 0.795&0.216 \\
         w/o static & 26.20& 0.769 & \underline{0.177} \\
        \midrule
        Full (Ours) & \textbf{27.30} & \textbf{0.816} & \textbf{0.130} \\
        \bottomrule
        % \multicolumn{4}{c}{(b)}
    \end{tabular}}
    \label{tab:ablation study}
\end{table}
% \begin{table}[]
%     \centering
%     \caption{\textbf{Ablation study.} The best performance is \textbf{boldfaced}, and the second is \underline{underlined}.}
%     \small
%     \setlength{\tabcolsep}{5.0pt}
%     \renewcommand\arraystretch{0.8}
%     \begin{tabular}{@{}lccc@{}}
%         \toprule
%         Methods & PSNR$\uparrow$ & SSIM$\uparrow$ & LPIPS$\downarrow$ \\
%         \midrule
%          w/o layered volume & 27.11& \underline{0.811} & 0.211\\
%         w/o optimized kernel &\underline{27.25}& 0.795&0.216 \\
%          w/o static & 26.20& 0.769 & \underline{0.177} \\
%         \midrule
%         Full (Ours) & \textbf{27.30} & \textbf{0.816} & \textbf{0.130} \\
%         \bottomrule
%     \end{tabular}
%     \label{tab:ablation study}
% \end{table}

%% file: sec/5_conclusion.tex
% \clearpage\mbox{}Page \thepage\ of the manuscript. This is the last page.
% Now we have reached the maximum length of an ECCV \ECCVyear{} submission (excluding references).
% References should start immediately after the main text, but can continue past p.\ 14 if needed.
\section{Conclusion}
We present \ourmethod, a novel framework for all-in-focus dynamic novel view synthesis from defocused monocular videos.  
To tackle defocus blur from video capture, we integrate the blur model into NeRF training. Particularly, we connect DoF rendering with volume rendering, proposing layered DoF volume rendering for dynamic NeRF. We modify the layered kernel to the ray-based kernel and apply an optimized sparse kernel to gather the rays. 
We conduct experiments to show our method can recover from defocus blur and produce satisfying results. 
Although this framework works well overall, some limitations remain, such as the inability to handle extreme defocus blur.
% The pre-trained optical flow network may produce unsatisfying details, and sometimes COLMAP may fail to estimate the camera poses.
% More details are in supplementary materials.
We plan to address these issues in future work. 
We declare to release the source code upon acceptance of the paper.

\clearpage  % TODO REVIEW/FINAL: This \clearpage needs to be removed from both review and camera-ready versions.

%% file: sec/supp.tex
% \clearpage
% \setcounter{page}{1}
% \maketitlesupplementary
This document includes the following contents:
\begin{enumerate}
    \item Discussion of more DoF-aware works.
    \item Layered composition for DoF rendering.
    \item Details of the dataset
    \item Failure case.
    \item More experiment results, including the results on the real defocused scene, quantitative results on individual scenes, comparison with state-of-the-art methods, and ablation study.
\end{enumerate}
% \begin{figure}
%     \centering
%     \includegraphics[width=0.76\linewidth]{images/kernel_new.pdf}
%     \caption{\textbf{Principle of layered DoF rendering.} 
%     }
%     \label{fig:kernel}
% \end{figure}
% \begin{figure}[t]
%     \centering
%     \includegraphics[width=0.95\linewidth]{images/failure.pdf}
%     \caption{\textbf{The failure case.} When the input views have severe defocus blur, the novel views may be unable to recover sharp details. 
%     }
%     \label{fig:failure}
% \end{figure}

\section{Discussion of More DoF-aware works}
Here we discuss the distinctions between our pipeline and several more DoF-aware works. The inputs of HDR-NeRF~\cite{mildenhall2021nerf} and NeRFocus~\cite{wang2022nerfocus} are all-in-focus sharp static images, and HDR-NeRF synthesizes HDR novel views with DoF effects, NeRFocus also generates defocused novel views. Our work, on the other hand, generates sharp novel views from defocused dynamic videos.

\cite{kaneko2022ar} is an unsupervised generative model, the training images are \textbf{single} images. \cite{kaneko2022ar} focuses on the unsupervised generative task and the generated images do not correspond with inputs. Ours is a supervised novel view synthesis task with defocused videos.

% LensNeRF\cite{kim2024lensnerf} restores a sharp NeRF from defocused \textbf{static} scene and also creates defocused views from in-focus \textbf{static} scene. Our task focuses on \textbf{dynamic} defocused video input. The technique of LensNeRF is valid for static scenes but does not model the temporal photometric consistency between video frames.

\cite{huang2023inverting} does not explore NeRF task, it only uses an implicit model to generate DoF, and the inputs are static image stack.

% In all, these tasks are different from our task: restoring a sharp dynamic NeRF from defocused videos.

\section{DoF Rendering}
We further describe the principles and technical details of DoF rendering. The layered composition in DoF rendering is similar to the multiplane image~\cite{tucker2020single} (MPI). 
The RGB image $C$ with its visibility weight $W$ is divided into layers by depth discretization. For current layered rendering~\cite{busam2019sterefo,zhang2019synthetic}, the composition weight $W$ is predefined by a certain fixed algorithm. On the other hand, in our method, we define the visibility weight $W$ from its physical principle, utilizing NeRF volume rendering to learn the compositing formation. 
% As shown in Fig.~\ref{fig:kernel}, the RGB image $C$ with its visibility weight $W$ is divided into layers by depth discretization. For current layered rendering~\cite{busam2019sterefo,zhang2019synthetic}, the composition weight $W$ is predefined by a certain algorithm. On the other hand, in our method, we define the visibility weight $W$ from its physical principle, utilizing NeRF volume rendering to learn the compositing formation. 

\section{The Dataset}
We conduct our experiments on dynamic scenes from a stereo dataset VDW~\cite{wang2023neural}. 
The dataset is collected from four data sources: movies, animations, documentaries, and web videos. The image sequences are over 1080p and are cropped at a resolution of $1880\times720$ or $1880\times800$.
Since the dataset provides RGB sequences with their corresponding aligned disparity sequences, we generate defocus blur from the state-of-the-art DoF rendering method~\cite{peng2022bokehme}. This method is proven to have better performance than current classical rendering and neural rendering methods. To be as realistic as real-world video-capturing scenarios, we adjust the focal distance along the scene disparity like the typical focusing. 

Although the VDW dataset is a large-scale dataset, not all scenes are suitable for our task due to two reasons: (1) many scenes lack moving objects and are static; (2) a large number of scenes exhibit minimal camera motion, resulting in insufficient parallax to obtain camera parameters. Therefore, we carefully select $8$ dynamic scenes from the dataset that are eligible for dynamic NeRF methods. 
These scenes consist of diverse object movements such as moving cars, walking, and opening a gate. The $8$ scenes are named Camp, Shop, Car, Mountain, Dining1, Dining2, Dock, Gate.
% We employ COLMAP to acquire camera parameters from the image sequences and downsample the image resolution to $940\times360$ or $940\times400$ for our experiments. 
We use the defocused sequences as inputs for DPT~\cite{ranftl2021vision} to obtain depth maps, and we utilize RAFT~\cite{teed2020raft} to generate optical flows, we also employ an instance segmentation network (Mask R-CNN~\cite{he2017mask}) to obtain motion masks for moving objects.

\begin{figure}[th]
    \centering
    \includegraphics[width=0.95\linewidth]{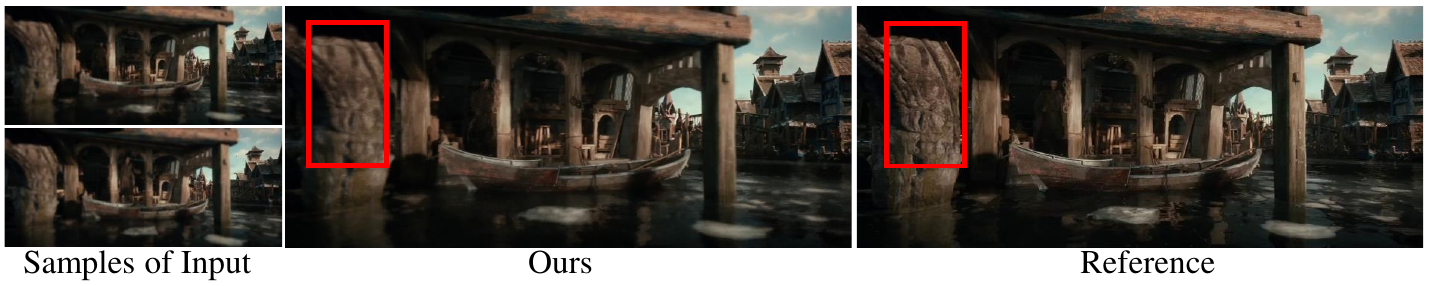}
    \caption{\textbf{The failure case.} When the input views have severe defocus blur, the novel views may be unable to recover sharp details. 
    }
    \label{fig:failure}
\end{figure}

\section{Failure Case}
As shown in Fig.~\ref{fig:failure}, our method may be unable to recover from extreme defocus blur. Extreme blur occurs when the focal distance is too close or too far from the camera, increasing the blur amount of the out-of-focus areas. The extreme defocus blur poses challenges for our method in two aspects: (1) the excessive information loss of the defocused regions may hinder the ability of the depth and flow prediction networks, perturbing the representation of dynamic scenes; (2) the model may get stuck in local minima reconstructing the scene when supervised by extreme blur. 
Furthermore, our model may fail to reconstruct sharp NeRF when all the frames of the sequences have consistent defocus blur, which means a certain object keeps focused for the whole sequence and other objects are consistently out-of-focus. However, capturing a dynamic scene often results in a shift in focal distance, so consistent blur is not common.
We require a larger memory cost than existing methods due to DoF modeling, but the additional cost is acceptable (Tab.~\ref{tab:cost}).
For future work, we aim to explore the integration of explicit representations to better utilize the in-focus regions from input frames and address the extreme blur issue.
\begin{table}[!h] 
\caption{Calculation cost.}
\scriptsize
\centering
\setlength{\tabcolsep}{1.0pt}
    \begin{tabular}{cccccc}
    \hline
    % & NSFF & HyperNeRF & DVS&RoDynRF&Ours (w/o DoF)&Ours\\ \hline
    & NSFF & HyperNeRF & DVS&RoDynRF&Ours\\ \hline
    Memory (G) & 11.9 & 14.9 & 12.4&11.9&20.1 \\
    % Test (G) & 2 & 0. & 0. \\ 
    \hline
    \end{tabular}
    % }
\label{tab:cost}
\end{table} 

\section{Experiments}
\begin{figure*}[h]
% \vspace{-10pt}
  \centering
  % \fbox{\rule{0pt}{0.5in} \rule{0.9\linewidth}{0pt}}
  \includegraphics[width=1.0\linewidth]{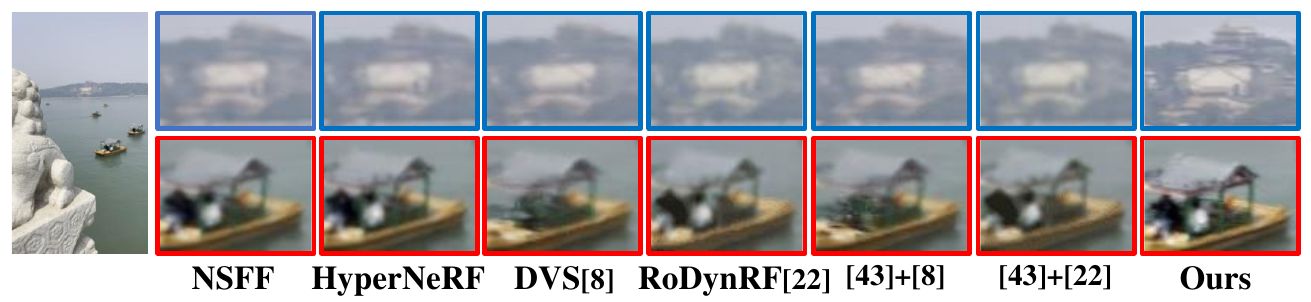}
  \vspace{-20pt}
   \caption{Results on real defocused video.}
   \label{fig:real}
\end{figure*}

\begin{figure}
    \centering
    \includegraphics[width=0.9\linewidth]{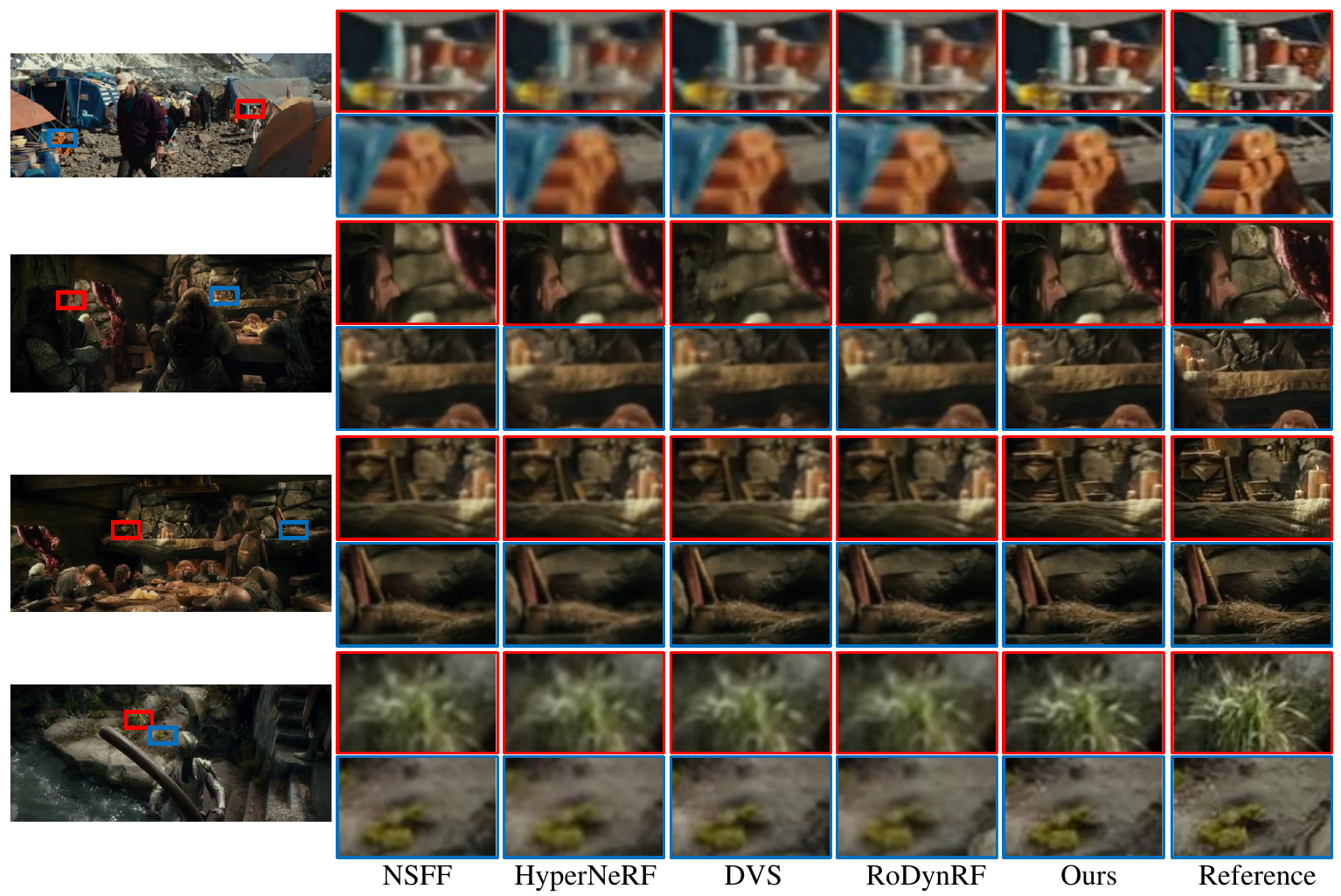}
    \caption{\textbf{The qualitative results with all dynamic NeRF baselines.} Compared with existing dynamic NeRF methods, our method generates sharper novel views that are more faithful and have more details. The scenes are Camp, Dining1, Dining2, Gate.} 
    \label{fig:compare_sota1}

\end{figure}

\begin{figure}
    \centering
    \includegraphics[width=0.9\linewidth]{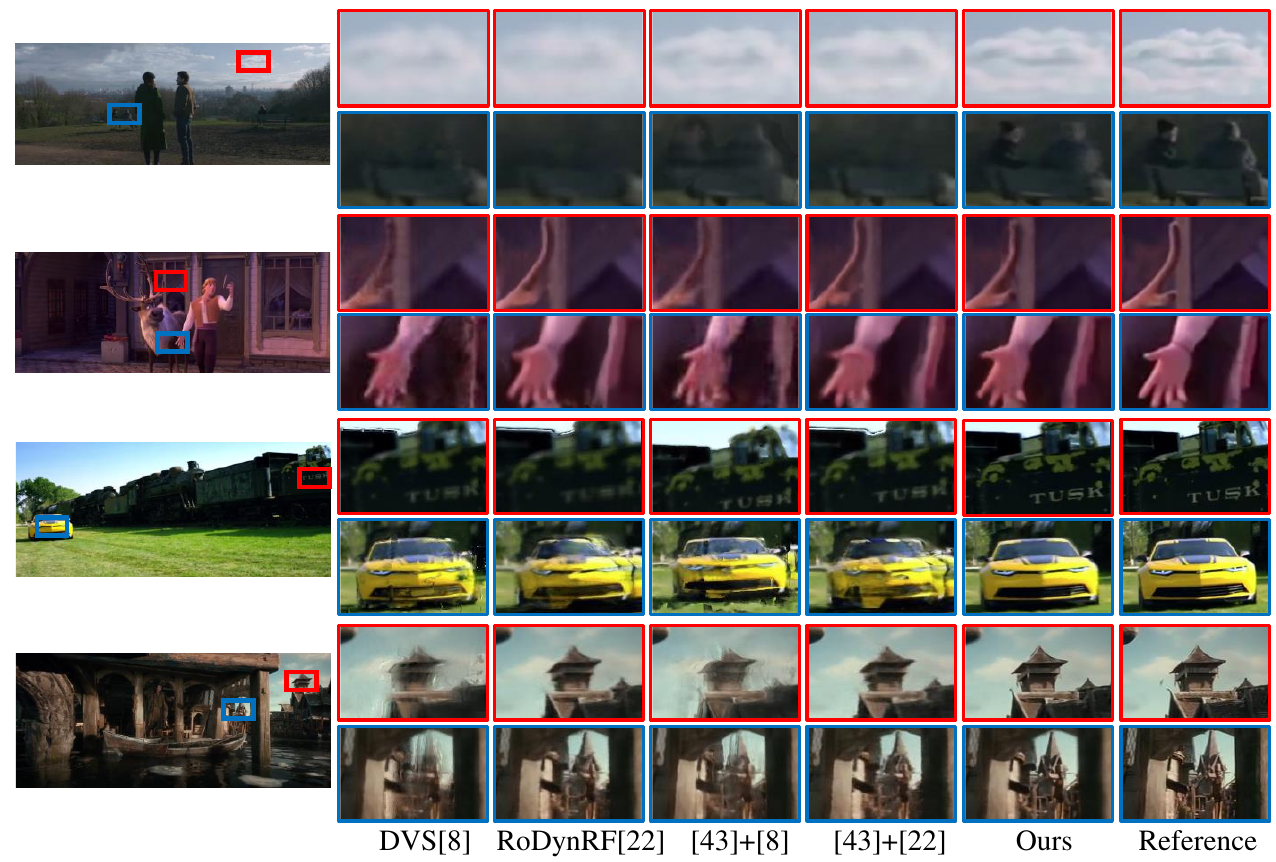}
    \caption{\textbf{The qualitative results with dynamic NeRF and their corresponding 2D image deblurring baselines.} The scenes are Mountain, Shop, Car, Dock.
    }
    \label{fig:compare_sota2}
\end{figure}

\begin{figure}
    \centering
    \includegraphics[width=0.9\linewidth]{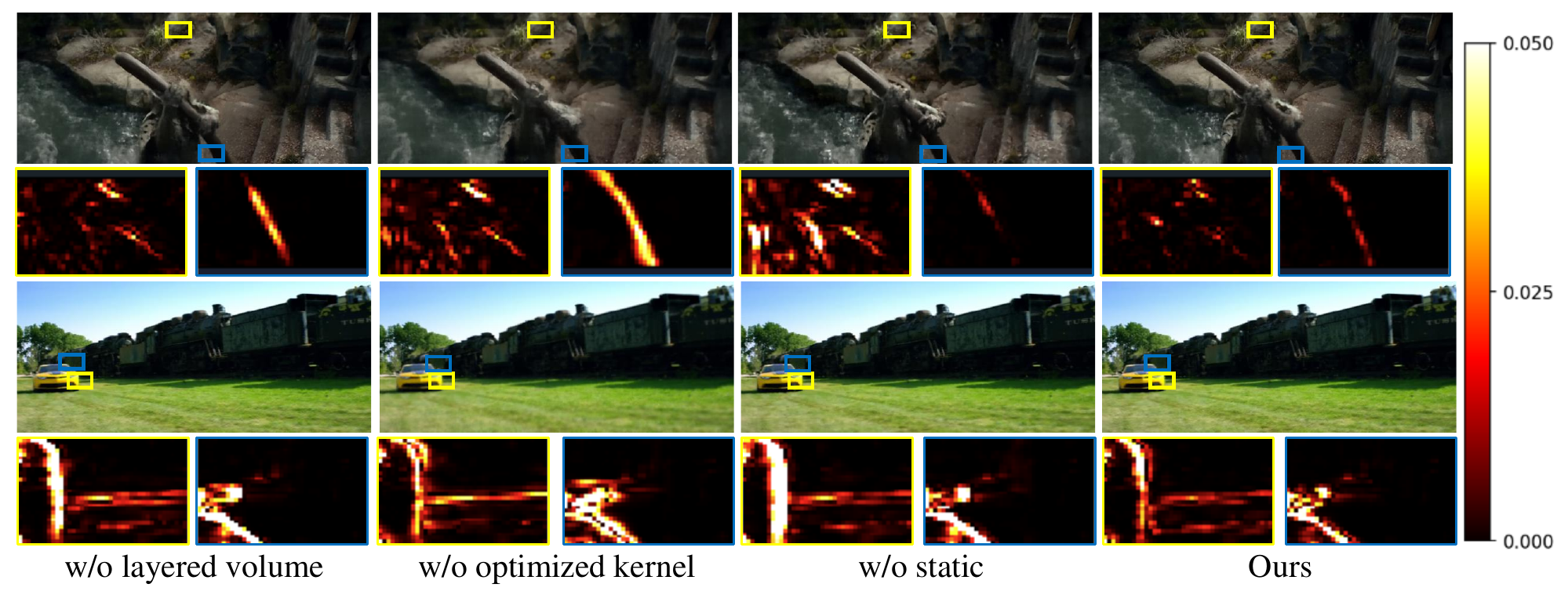}
    \caption{\textbf{The visualizations of the ablation study.} The corresponding error map is visualized at the bottom, where darker regions indicate smaller errors. We define the error range from $0$ to $0.05$. Our full model has the smallest error overall.
    }
    \label{fig:ablation_supp}
\end{figure}

We show the quantitative results with state-of-the-art methods on all scenes in Tab.~\ref{tab:baseline_nvds}. We also collect one real dynamic defocused video for evaluation. As shown in Fig.~\ref{fig:real}, our method is more stable and generates more reliable sharp details.
The ablation results are in Tab.~\ref{tab:ablation_dynamic} and Tab.~\ref{tab:ablation_full}
We show qualitative comparisons with state-of-the-art methods in~\cref{fig:compare_sota1,fig:compare_sota2}. We choose different scenes from the main article to show results on all scenes.
We visualize the ablation results in Fig.~\ref{fig:ablation_supp}. 
% Our method with the full module is proven to be more stable than other baselines.
% \vspace{-10pt}

\begin{table*}[htp]
	\centering
	\caption{Quantitative results on the synthesized dataset. The best performance is in \textbf{boldface}, and the second best is \underline{underlined}. 
 Although the baselines achieve better results on a certain indicator in some scenes, our method achieves better visualization quality and restores sharper details than other baselines as shown in the qualitative results and the supplementary video.
 % Although for some scenes current baselines achieve better results on a certain indicator, these baselines are not stable across all scenes. We achieve better quality and restore sharper details than other baselines in the qualitative results and the supplementary video where \ourmethod achieves better multi-view consistency and restores sharper details than other baselines.
 } 
	\resizebox{1.0\linewidth}{!}
	{
    \setlength{\tabcolsep}{3pt}
	\renewcommand\arraystretch{1.0}
	\begin{NiceTabular}{l|ccc|ccc|ccc|ccc}
		\toprule
		\multicolumn{1}{l}{\multirow{2}{*}[-0.5ex]{Method}} &
  \multicolumn{3}{c}{Camp} & \multicolumn{3}{c}{Shop}& \multicolumn{3}{c}{Car}& \multicolumn{3}{c}{Mountain} \\
		\cmidrule{2-13}
		~ &  PSNR$\uparrow$ &  SSIM$\uparrow$&LPIPS$\downarrow$ & PSNR$\uparrow$ & SSIM$\uparrow$ & LPIPS$\downarrow$ &PSNR$\uparrow$ &  SSIM$\uparrow$ & LPIPS$\downarrow$ &PSNR$\uparrow$ &  SSIM$\uparrow$&LPIPS$\downarrow$ \\
        \midrule
		\midrule
		DVS~\cite{gao2021dynamic}& 17.18& 0.525&0.399& 27.05 & 0.842&0.216 & 24.57& 0.795& 0.240&\textbf{33.54}&\underline{0.896}&0.174\\
		NSFF~\cite{li2021neural}& \textbf{21.17} & \underline{0.643} &  0.310& 26.54 &0.838 & 0.207& \underline{25.57}&\underline{0.799}&0.223&32.11&0.891&0.158\\
        HyperNeRF~\cite{park2021hypernerf}& \underline{21.08} & 0.627 &\underline{0.294}& 27.21 &0.834 & 0.216& 25.44 &0.769&0.216&\underline{33.02}&0.886&0.160\\
        RoDynRF~\cite{liu2023robust}& 20.99 & 0.597&0.312& \textbf{28.53}& 0.844& 0.213 &22.62 & 0.726 & 0.264&28.71&0.858&0.194\\
        \cite{son2021single} + DVS& 17.07 & 0.531&0.352& 26.47& 0.850& \underline{0.155} &23.72 & 0.786 & \underline{0.171}&31.87&\textbf{0.900}&\underline{0.128}\\
        \cite{son2021single} + RoDynRF&20.53 & 0.564& 0.345 & \underline{28.15}& \textbf{0.856}&0.159 &22.65&0.756&0.197&27.46&0.863&0.149\\
		\ourmethod (Ours)& 20.73 & \textbf{0.644}&\textbf{0.207} & 27.01& \underline{0.854} &\textbf{0.117}& \textbf{26.79}& \textbf{0.852} & \textbf{0.123}& 32.44&\textbf{0.900}&\textbf{0.079}\\
        \midrule
        \multicolumn{1}{l}{\multirow{2}{*}[-0.5ex]{Method}} &
  \multicolumn{3}{c}{Dining1} & \multicolumn{3}{c}{Dining2}& \multicolumn{3}{c}{Dock}& \multicolumn{3}{c}{Gate} \\
		\cmidrule{2-13}
  ~ &  PSNR$\uparrow$ &  SSIM$\uparrow$&LPIPS$\downarrow$ & PSNR$\uparrow$ & SSIM$\uparrow$ & LPIPS$\downarrow$ &PSNR$\uparrow$ &  SSIM$\uparrow$ & LPIPS$\downarrow$ &PSNR$\uparrow$ &  SSIM$\uparrow$&LPIPS$\downarrow$ \\
        \midrule
		\midrule
        DVS~\cite{gao2021dynamic}& 23.07 & 0.712&0.287&27.85 & \underline{0.822}&0.165 & 27.54& 0.807& 0.220&22.60&0.713&0.231\\
		NSFF~\cite{li2021neural}& \underline{30.54} & \underline{0.865} &  \underline{0.175}& 27.54 &0.812 & \underline{0.152}& \textbf{28.42}&\underline{0.822}&0.216&\underline{24.17}&\underline{0.754}&0.228\\
        HyperNeRF~\cite{park2021hypernerf}& 30.48 & 0.854 &0.198& 27.55 &0.818 & 0.165& 27.94 &0.806&0.212&22.99&0.647&0.202\\
        RoDynRF~\cite{liu2023robust}& 29.21 & 0.803&0.243&\textbf{28.25} &0.816 & 0.171 &27.07 & 0.789 & 0.214&24.06&0.726&0.201\\
        \cite{son2021single} + DVS& 22.91 & 0.724&0.252 &27.78 & \textbf{0.826}& 0.153 &26.94 & 0.790 & 0.189&19.36&0.647&0.265\\
        \cite{son2021single} + RoDynRF&28.96 &0.812  &0.205  & \underline{27.92}& 0.815& 0.156& 26.83&0.797&\underline{0.178}&23.85&0.741&\textbf{0.182}\\
		\ourmethod (Ours)& \textbf{31.05} & \textbf{0.877}&\textbf{0.105} &27.84 & 0.817&\textbf{0.094}& \underline{28.12}& \textbf{0.823} & \textbf{0.131}& \textbf{24.41}&\textbf{0.757}&\underline{0.185}\\
    \bottomrule
	\end{NiceTabular}
	}\label{tab:baseline_nvds}
\end{table*}

\begin{table*}[h]
	\centering
	\caption{Ablation results on the synthesized dataset. The results only calculate the dynamic regions from motion masks.
	The best performance is in \textbf{boldface}, and the second best is \underline{underlined}. The results show that our method works best with all the modules, and missing one of them causes performance degradation.} 
	\resizebox{1.0\linewidth}{!}
	{
    \setlength{\tabcolsep}{3pt}
	\renewcommand\arraystretch{1.0}
	\begin{NiceTabular}{l|ccc|ccc|ccc|ccc}
		\toprule
		\multicolumn{1}{l}{\multirow{2}{*}[-0.5ex]{Method}} &
  \multicolumn{3}{c}{Camp} & \multicolumn{3}{c}{Shop}& \multicolumn{3}{c}{Car}& \multicolumn{3}{c}{Mountain} \\
		\cmidrule{2-13}
		~ &  PSNR$\uparrow$ &  SSIM$\uparrow$&LPIPS$\downarrow$ & PSNR$\uparrow$ & SSIM$\uparrow$ & LPIPS$\downarrow$ &PSNR$\uparrow$ &  SSIM$\uparrow$ & LPIPS$\downarrow$ &PSNR$\uparrow$ &  SSIM$\uparrow$&LPIPS$\downarrow$ \\
        \midrule
		\midrule
		w/o layered volume& \underline{20.89}&\underline{0.664}& \underline{0.245}& 19.65&0.583&0.245 &24.34&0.886&\underline{0.077}&\textbf{26.09}&\underline{0.818}&\underline{0.125}\\
        w/o optimized kernel & \textbf{20.91} & 0.641 & 0.331&\underline{19.76}&\underline{0.592}&0.256&\underline{25.04}&\underline{0.888}&0.130&25.90&0.796&0.196 \\
         w/o static & 20.68& 0.658 & 0.252&19.42&0.567&\underline{0.221}&22.96&0.867&0.085&25.31&0.780&\underline{0.125} \\
        % \midrule
         Full (Ours) & 20.70 & \textbf{0.665} & \textbf{0.220}&\textbf{19.93}&\textbf{0.595}&\textbf{0.203}&\textbf{25.18}&\textbf{0.901}&\textbf{0.075}&\underline{25.95}&\textbf{0.820}&\textbf{0.106}\\
        \midrule
        \multicolumn{1}{l}{\multirow{2}{*}[-0.5ex]{Method}} &
  \multicolumn{3}{c}{Dining1} & \multicolumn{3}{c}{Dining2}& \multicolumn{3}{c}{Dock}& \multicolumn{3}{c}{Gate} \\
		\cmidrule{2-13}
  ~ &  PSNR$\uparrow$ &  SSIM$\uparrow
  $&LPIPS$\downarrow$ & PSNR$\uparrow$ & SSIM$\uparrow$ & LPIPS$\downarrow$ &PSNR$\uparrow$ &  SSIM$\uparrow$ & LPIPS$\downarrow$ &PSNR$\uparrow$ &  SSIM$\uparrow$&LPIPS$\downarrow$ \\
        \midrule
		w/o layered volume&\underline{30.99}&\underline{0.866}&\underline{0.099}&25.29&0.724&\textbf{0.135}&27.40&\textbf{0.779}&\textbf{0.136}&20.89&0.594&\underline{0.294}\\
        w/o optimized kernel &30.61&0.851&0.144&\underline{25.66}&\underline{0.733}&0.221&\underline{27.47}&0.767&0.219&\underline{20.97}&\textbf{0.615}&0.472\\
         w/o static &30.59&0.850&0.141 &24.69 & 0.697 & 0.144&\textbf{27.65}&\underline{0.775}&0.180 &20.19&0.558&\textbf{0.280}\\
        % \midrule
         Full (Ours) &\textbf{31.01}&\textbf{0.871}&\textbf{0.085}& \textbf{25.76} & \textbf{0.745}&\underline{0.138} & 27.24&0.774&\underline{0.150}&\textbf{21.03}&\underline{0.603}&0.316 \\
        \bottomrule
	\end{NiceTabular}
	}\label{tab:ablation_dynamic}
\end{table*}

\newpage

% \newpage
 
\begin{table*}[h]
	\centering
	\caption{Ablation results on the synthesized dataset. The results calculate the whole image.
	The best performance is in \textbf{boldface}, and the second best is \underline{underlined}. 
     The results show that our method works best with all the modules, and missing one of them causes performance degradation.
 % Although for some scenes current baselines achieve better results on a certain indicator, these baselines are not stable across all scenes, and we urge readers to view the visualizations where \ourmethod achieves better quality.
 } 
	\resizebox{1.0\linewidth}{!}
	{
    \setlength{\tabcolsep}{3pt}
	\renewcommand\arraystretch{1.0}
	\begin{NiceTabular}{l|ccc|ccc|ccc|ccc}
		\toprule
		\multicolumn{1}{l}{\multirow{2}{*}[-0.5ex]{Method}} &
  \multicolumn{3}{c}{Camp} & \multicolumn{3}{c}{Shop}& \multicolumn{3}{c}{Car}& \multicolumn{3}{c}{Mountain} \\
		\cmidrule{2-13}
		~ &  PSNR$\uparrow$ &  SSIM$\uparrow$&LPIPS$\downarrow$ & PSNR$\uparrow$ & SSIM$\uparrow$ & LPIPS$\downarrow$ &PSNR$\uparrow$ &  SSIM$\uparrow$ & LPIPS$\downarrow$ &PSNR$\uparrow$ &  SSIM$\uparrow$&LPIPS$\downarrow$ \\
        \midrule
		\midrule
		w/o layered volume& \textbf{20.90}& 0.641& 0.230&26.78&\textbf{0.854}&0.154&25.93&0.835&0.134&\textbf{32.57}&0.899&0.099 \\
        w/o optimized kernel & 20.86 & 0.615 & 0.317&26.85&\underline{0.853}&0.180&\textbf{26.72}&0.810&0.218&31.86&0.881&0.157 \\
         w/o static & \underline{20.77} & 0.633 & 0.238&26.33&0.825&0.180&24.69&0.798&0.149&31.02&0.862&0.117 \\
        % \midrule
         Full (Ours) & 20.73 & \textbf{0.644} & \textbf{0.207}&\textbf{27.02}&\textbf{0.854}&\textbf{0.115}&\underline{26.79}&\textbf{0.852}&\textbf{0.123}&\underline{32.44}&\textbf{0.900}&\textbf{0.079} \\
        \midrule
        \multicolumn{1}{l}{\multirow{2}{*}[-0.5ex]{Method}} &
  \multicolumn{3}{c}{Dining1} & \multicolumn{3}{c}{Dining2}& \multicolumn{3}{c}{Dock}& \multicolumn{3}{c}{Gate} \\
		\cmidrule{2-13}
  ~ &  PSNR$\uparrow$ &  SSIM$\uparrow$&LPIPS$\downarrow$ & PSNR$\uparrow$ & SSIM$\uparrow$ & LPIPS$\downarrow$ &PSNR$\uparrow$ &  SSIM$\uparrow$ & LPIPS$\downarrow$ &PSNR$\uparrow$ &  SSIM$\uparrow$&LPIPS$\downarrow$ \\
        \midrule
		w/o layered volume& \underline{30.94}& \underline{0.872}& \underline{0.118}&27.45&0.806&\underline{0.098}&\underline{28.24}&\underline{0.833}&\textbf{0.118}&\underline{24.26}&\underline{0.749}&\underline{0.189} \\
        w/o optimized kernel & 30.64 & 0.857& 0.152&\textbf{28.33}&\textbf{0.830}&0.144&\textbf{28.80}&\textbf{0.834}&\textbf{0.118}&23.93&0.681&0.438 \\
         w/o static & 30.31 & 0.849 & 0.161&25.50&0.734&0.128&27.87&0.804&0.183&23.14&0.645&0.259 \\
        % \midrule
         Full (Ours) & \textbf{31.05} & \textbf{0.877} & \textbf{0.105}&\underline{27.85}&\underline{0.817}&\textbf{0.094}&28.12&0.823&\underline{0.131}&\textbf{24.41}&\textbf{0.757}&\textbf{0.185} \\
        \bottomrule
	\end{NiceTabular}
	}\label{tab:ablation_full}
\end{table*}

\newpage